# Identifying Human Indoor Daily Life Behavior employing Thermal Sensor Arrays (TSAs)

**Dina E. Abdelaleem[1, 2*†], Hassan M. Ahmed[3†], M. Sami Soliman[4, 5], Tarek M. Said[6]**

[1]Fayoum University, Department of Electrical Engineering, Fayoum, Egypt.
[2]Nahda University, Department of Communications and Computers, Beni-Suef, Egypt
[3]AMI-Lab, Département d'Informatique, Faculté des sciences, Université de Sherbrooke
[4]Nahda University, Department of Communications and Computers, Beni-Suef, Egypt
[5]Seismology Department, National Research Institute of Astronomy and Geophysics, Cairo, Egypt
[6]Fayoum University, Department of Electrical Engineering, Fayoum, 6351, Egypt

*Corresponding author: Dina E. Abdelaleem (e-mail: dina.abdelaleem@nub.edu.eg).

† These authors have equally contributed to the manuscript.

ABSTRACT

Daily activity monitoring systems used in households provide vital information for health status, particularly with aging residents. Multiple approaches have been introduced to achieve such goals, typically obtrusive and non-obtrusive. Amongst the obtrusive approaches are the wearable devices, and among the non-obtrusive approaches are the movement detection systems, including motion sensors and thermal sensor arrays (TSAs). TSA systems are advantageous when preserving a person's privacy and picking his precise spatial location. In this study, human daily living activities were monitored day and night, constructing the corresponding activity time series and spatial probability distribution and employing a TSA system. The monitored activities are classified into two categories: sleeping and daily activity.

Results showed the possibility of distinguishing between classes regardless of day and night. The obtained sleep activity duration was compared with previous research using the same raw data. Results showed that the duration of sleep activity, on average, was 9 hours/day, and daily life activity was 7 hours/day. The person's spatial probability distribution was determined using the bivariate distribution for the monitored location. In conclusion, the results showed that sleeping activity was dominant. Our study showed that TSAs were the optimum choice when monitoring human activity. Our proposed approach tackled limitations encountered by previous human activity monitoring systems, such as preserving human privacy while knowing his precise spatial location.

I. INTRODUCTION

Understanding human behavior is essential for society since it can be used to spot trends, provide rationales for people's actions, and much more. The more human behavior and activity are comprehended, the better explanations can be provided for why that individual behaved in a particular manner in various circumstances. Human behavior refers to the activities or responses a person makes in response to an internal or external stimuli condition [1]. Understanding the driving force behind someone's behavior helps us better comprehend him. The Internet of Things, or IoT, has garnered more attention recently and has been used in many applications, particularly in the healthcare industry. Different medical IoT devices help older people adapt and live alone to overcome the problem of the increasing cost of seeking long-term care [2]–[4]. Medical IoT devices encompass different sensing technologies, such as wearable-based sensors, vision-based sensors, and ambient sensing sensors.

Continuous wear or carrying of a device is generally required for wearable-based sensors. However, they are inconvenient for older people and are considerably more complicated to handle for older persons who have dementia [5], such as Alzheimer's, because there is an excellent likelihood that they may forget to carry these devices. On the other hand, there are still unresolved issues with wearable device design, including energy efficiency, manufacturing, and lightweight [6].

Vision-based sensing systems comprise several video cameras connected to a computer machine running software to process the captured images in real time. In real-world situations, these systems operate incredibly well, although they don't preserve the privacy of the individuals [7]. Ambient sensing tools like passive infrared (PIR) and motion sensors are utilized in a home environment. These gadgets protect privacy but do not function effectively in multioccupancy home environments [5].

More focus has recently been placed on optimizing the performance and privacy in applications for residential environments [8]–[10]. The motion sensors discover whether there is a movement in the field of view (FOV) of the sensor, and these sensors effectively detect whether there is an activity in their vicinity. However, they fail to count the number of people inside the place where the monitoring process is performed. Moreover, it fails to determine the spatial location of the activity done. It does not differentiate between the person standing, sitting, or lying down. On the contrary, TSA overcomes the previously mentioned limitations of wearable and vision-based sensors.

It is better to use TSA-type sensors, as they are characterized by preserving privacy, noncontact capabilities, and low cost [11]. The low price, privacy-preserving, and noncontact features of TSA sensors make them viable new sensing technologies for applications focused on people. The TSA captures a full image of the room where the person is to be monitored, where this image is captured every specific time unit and is stored and processed [12], [13]. The thermal sensor determines the spatial location of the individual, which helps count the number of individuals in the monitored place. A thermal sensor helps classify a person's activities at night and day. In this study, activities were categorized into two activities: daily Activity and sleeping Activity.

The acquired data from such sensors, the TSA sensors, can be visualized/reconstructed as either a set of single frames/images or a stack of frames composing a video. Such a video included several objects, such as a single individual or a group of individuals doing a particular activity or hot objects inside the room. The activity was defined by its duration and spatial location for the person's object.

Identifying and detecting objects in such images or video is a task in the computer vision field known as object detection. Object detection can be determined manually and automatically.

There were limitations to counting the time of a specific activity manually in case of monitoring multiple individuals, and in case of massive data, these limitations were (1) manual object detection being ineffective, (2) inaccuracy of the human vision, and (3) time-consuming. On the other hand, automatic object detection had several limitations; selecting a suitable threshold value was challenging.

A TSA-based system was proposed that monitored the daily life activities of individuals inside their residences. Moreover, an algorithm was proposed to automate the activity detection and calculation for the individual inside his residency; where this algorithm detected the individual's location inside the captured image by the TSA and classified the type of activity performed by the individual, leading to calculating the total time of such activity.

This research focused on monitoring sleep activity and calculating the number of hours of sleep activity from the 7th of April to the 6th of May 2021 for the person to be monitored. The rest of this article was divided into sections: Section II summarizes the related work. Section III explains the methodology. Results are presented in Section IV. Discussion of results was presented in Section V. Finally, the conclusion is given in Section VI.

## II. RELATED WORK

Various sensing techniques have been investigated in research to determine the optimum approach to determine the total number of individuals living in a given environment for various objectives [14], [15]. It was necessary to estimate the number of individuals occupying the building to determine if persons are active or inactive and if there is a distance between people and each other. Chen *et al.* [14] presented a comprehensive review for determining occupancy and calculating the number of people inside a building. In addition, Chen *et al.* presented a review of various systems that use sensor fusion to identify building occupancy and a comparison of the used sensors. Saha *et al.* [15] proposed a comprehensive assessment of the many techniques used to identify building occupancy, count people, and monitor those people.

However, PIR sensors suffer from the limitation of not differentiating whether there is a single individual or multiple individuals in the room. Howedi *et al.* [16] conducted research about the PIR sensor to distinguish between single and multioccupancy environments to evaluate the visit time of older persons in a single-inhabitant climate. They did this by assessing the unpredictability of the PIR-based binary data using various entropy measures.

Several researchers [17]–[20] have studied indoor multioccupancy by deploying PIR sensors in the environment. The studies mentioned above, however, could not offer a precise estimate of the population; instead, they could only determine whether several people were present in the area. They also placed much reliance on the sensor configuration.

Beltran *et al*. [21] suggested using a TSA and a PIR sensor to determine occupancy in a multi-modal system. The TSA used the PIR sensor to count the number of individuals in the environment and to detect an empty occupancy environment. However, because of the limitation of the PIR sensor, it was more likely to detect human radiation when the background temperature rose; the suggested system might not be able to calculate occupancy after someone had been still for an extended amount of time, such as when they were sleeping.

Utilizing low-resolution thermal sensor array (TSA) in indoor human-centric systems has recently gained more attention [22]–[25]. In human-centered applications, the thermal sensor array (TSA) has been used as a cutting-edge technology for sensing to minimize the gap between performance and privacy found in vision- and wearable-based techniques. As work presented by [26]–[28] showed, thermal sensors have recently been used to diagnose some temperature-related diseases, such as fever, especially with the global COVID-19 19.

As work presented by [29]–[32], Many applications, including occupancy detection, estimation, security, surveillance, traffic management, and activity recognition, have benefited from using thermal sensors. In the research conducted by G. Spasov *et al.* [29], MLX90640 is used to develop an IoT solution for security intelligent systems, where the primary motivation was to help older people adapt to the ambient. In addition, the paper discussed the privacy issues when using traditional cameras to recognize activity and human localization, which motivated the use of thermal sensors to solve the previously mentioned issues. Also, the research indicated that hot-moving objects other than the object to be monitored could cause a conflict in detecting and estimating occupancy. Compared with [29], the activity recognition was accomplished in our work, and the daily activity was classified into 2 categories. Health, household use, and system control are among the applications for thermal sensors [33], [34]. In addition to its use in the security field [35].

Similarly, sensor site adaptation issues were presented in other research [36]–[39]. Aloulou et al. [40] presented an adaptable method for motion sensor plug-and-play systems utilized for older people's ambient assistive living. According to the previous research reported in [8], [41] on estimating human distance and occupancy, An approach to sensor placement and fusion in an accessible position was suggested. Such an approach could recognize overlapping sensor areas from low-resolution thermal images to overcome the difficulty of allowing TSA to check a large area to detect the multioccupancy environment, human-to-human distance, and single-based TSA processing for human-to-sensor distance.

Abdallah Naser *et al*. [42] evaluated the occupancy in the interior environment and provided a semantic of human presence. Additionally, it could segregate human presence and count people in various sensor placements, interior settings, and at various sensor distances. The proposed framework was assessed using classification and regression machine learning techniques to predict occupancy at different sensor locations, the number of inhabitants, surroundings, and human distance. The accuracy of the classification strategy employing the adaptive boosting algorithm was demonstrated in this research to be 98.43% for vertical sensor sites and 100% for overhead sensor locations, respectively.

Hassan M. Ahmed and Bessam Abdulrazak [43] proposed an IoT system for monitoring an indoor activity of daily living (ADLs) using a thermal sensor array (TSA) in a room for a period of about 35 days. Their study investigated daily activity, sleeping activity, and no activity for an individual using a thermal sensor array (TSA). This work also introduced TSAs to estimate a person's average room temperature while maintaining the monitored individual's privacy. The person's activity was observed by tracking thermal image pixels associated with the activity's spatial locations within the room. The temporal temperature of the related pixels is measured to create various activity vectors/arrays for the person. These vectors were then transmitted to a cloud server annotated with the relevant timestamp to be stored and processed there. The test results exhibited the person's behavior for ten consecutive days in April 2021. The experiment demonstrated how accurately a person's spatial location indoors could be detected by TSA technology. Additionally, while respecting the subject's anonymity (privacy), the experiment allowed for the prediction of bathroom visits and the estimation of the subject's body temperature throughout the monitoring period. The same data they collected is being utilized, and a novel approach is being presented to detect the similar classes they detected. Moreover, they have concluded another research to study the availability of detecting bathroom related disorders such as the Urinary Tract Infection (UTI) [3]. Continuing with the same topic of utilizing the IoT in the field of medical support, the same team concluded another research to study the effect of an inhibiting drug on the human indoor activity by means of monitoring the indoor activity by IoT activity sensors [44].

The use of a thermal sensor to monitor human behavior had advantages, as presented in their study, that was summarized as preserving the privacy of participating objects, tracking the precise spatial location of the person without any contact, counting the number of objects, and classifying the activities of the object to be monitored.

Abdallah Naser *et al*. [45] proposed an approach for fusing several TSAs to cover a broad inspection region, allowing additional human-centered apps to run on a centralized cloud platform and locate areas of overlap between two or more low-resolution TSAs. This approach was characterized by improving the privacy of the participating objects to prevent a third party's construction of human images during data transfer and cloud storage, which is flexible enough to operate in different interior configurations and sensor locations. The proposed approach used machine learning techniques and use-case situations to deliver a 92.5% accuracy on average performance. This proposed approach was distinguished from other methods mentioned in that it covers an ample inspection space to avoid losing any information related to the object to be monitored.

### III. METHODOLOGY

The following steps were followed to conclude our methodology: (1) Hardware, (2) Data pre-processing, (3) Video frames construction, (4) Object detection and scatter diagrams construction, (5) Video frames construction, (6) Calculating activity time, and (7) Calculating spatial distribution.

A person was tracked for 35 days using TSA (MLX90640). Two raw data files were obtained; the timestamp in the first file and the sensor data in the second file were combined into a single data frame. The number of hours per day was calculated to calculate the missing data rate, and every day with more than 50% raw sensor data was considered, while every day with less than 50% raw sensor data was excluded (else). The thermal image was constructed, and object detection techniques were applied to determine the objects in the image. Activity time series were built to track and classify a person's activity and construct a spatial probability distribution.

The raw data was obtained from a thermal sensor array to monitor a person's activity. Twenty percent of the data were missing for the entire monitoring period, meaning seven days of the 35 days were lost. The days with data of more than or equal to 50% were used. A corresponding video frame was constructed from the raw data for each day. Then, image processing techniques were applied to each frame. Following that, object detection techniques were used for every frame. To follow the human object and exclude other heat sources across the video frames, all X-axis and Y-axis for all objects in a file and the axis of the fixed object (heater) were dropped. Then, the centroid for every object was placed in a scatter diagram, and the timestamp of the sleeping and daily activities was plotted to track the overall activity of the object being monitored. The software used was Python 3.8.8 in Jupiter Notebook, and the libraries used were Matplotlib, Pandas, NumPy, and OpenCV.

#### A. Hardware Setup and Data Acquisition

A thermal sensor array was used to continuously observe and record a male subject's behavior, as shown in Figure 1. The thermal sensor used is MLX90640, and unit processing is used to analyze the data obtained from the thermal sensor. The monitoring of one individual serves as a proof-of-concept and feasibility test. The person's activity was observed by tracking thermal image pixels associated with the activity's spatial locations within the room. The temporal temperature of the related pixels is measured to create various activity vectors/arrays for the person. These vectors were then transmitted to a cloud server annotated with the pertinent timestamp so they could be kept and examined there. The extracted raw data from the thermal sensor was categorized into two classes: daily activity and sleeping activity. The steps that were taken to monitor three classes of human behavior, such as sleeping Activity,

daily Activity, and no activity, are presented in

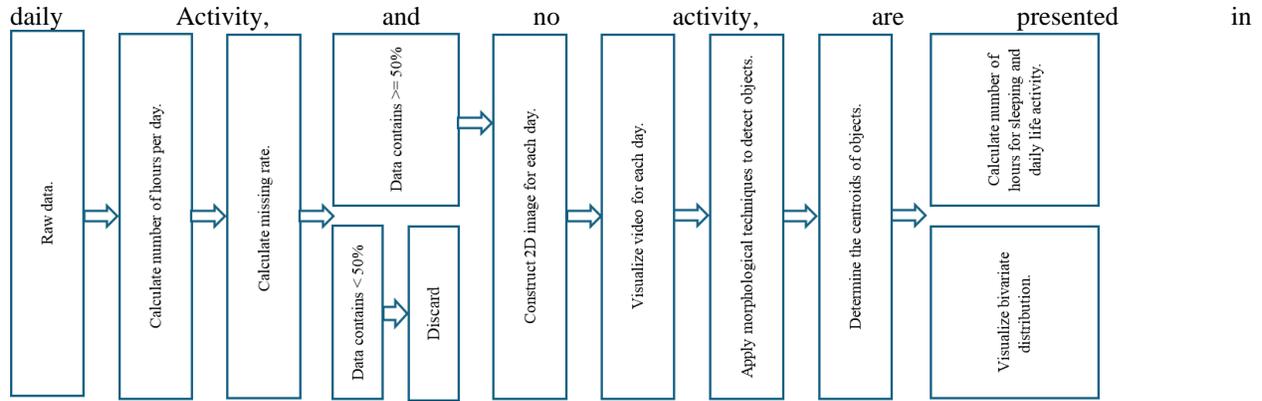

Figure 2.

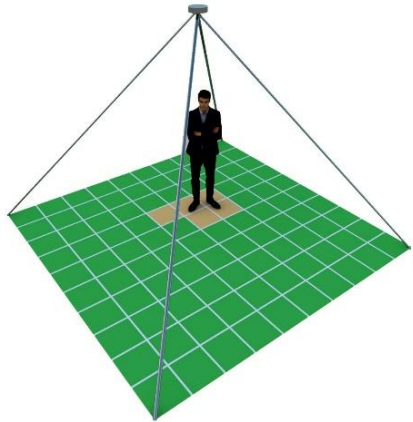

**Figure 1.** Representation of the sensor location on the ceiling and subjects in the room, along with their thermal distribution

### B. Data Pre-processing

The raw data obtained from the sensor was stored in two files: one with a timestamp and another with raw data from the sensor. The two files were combined into one file, and the timestamp and the sensor data were matched.

From the combined file, which contained the data of the sensor and timestamp, A vector for the number of hours was created, which contained data to calculate the number of hours per day. The raw data which was obtained from a thermal sensor included a time series for 35 days; the number of days which was missed data was calculated by calculating the number of missing data per day by calculating the number of days that contained data greater than or equal to 12 hours for each day and calculating the number of days which contained data of less than 12 hours for each day. The percentage of missing data per day was calculated after calculating the number of hours greater than or equal to 50% per day. However, the missing data can be imputed by different methods ranging from substitution methods such as hot-deck method to model based methods such as gaussian model based methods [46].

### C. Video Frames Construction

The raw data from the thermal sensor was in a 1D array, converted to a 2D array to construct a 2D image for each day, having greater than or equal to 50% of the data and visualizing each day as a video. The 2D image was constructed by a combination of the data frame of the time stamps data and the sensor raw data in one data frame; then, each row, which represented the capture of every minute from the thermal sensor array, was converted from 1D array to 2D array by splitting each row to 16 segments and getting 2D array 16*12. The 2D image for all days was constructed as shown in Figure 3.

### D. Object Detection and Scatter Diagram Construction

The objects were automatically detected in the image using morphological techniques, and their centroids were determined. After constructing a 2D image, the thermal image was converted to a level image; then the grey image was contoured, edge detection was applied for them, and then the centroids were detected for all objects in the image, as presented in Figure 4.

### E. Calculating Activity Time

For each frame, the centroid of the person was determined; then, the corresponding time index was obtained to get a temporal domain and put 1 in the time index. After tracking the centroids for all objects in the 2D image, the object whose behavior was to be monitored was tracked throughout the day to classify the two activities mentioned earlier. Then, sleeping and daily activities were detected each day for days greater than or equal to 50% of the data. The sleeping activity was calculated when the person was on the bed, and the daily activity was calculated when the person was on the table eating and working. The hours spent sleeping and doing daily activities were calculated each day.

### F. Calculating Spatial Distribution

Bivariate distribution was calculated to represent the spatial distribution of the person's activity. The spatial distribution was presented to show the person is in the location and showed the relationship between the person's activity and spatial location, so it was easy to show how the person spent his time in the area and classified his activity. Logarithmic normalization was applied to implement the bivariate distribution technique, as the axis was extensive; reducing the axis made the bivariate distribution more evident.

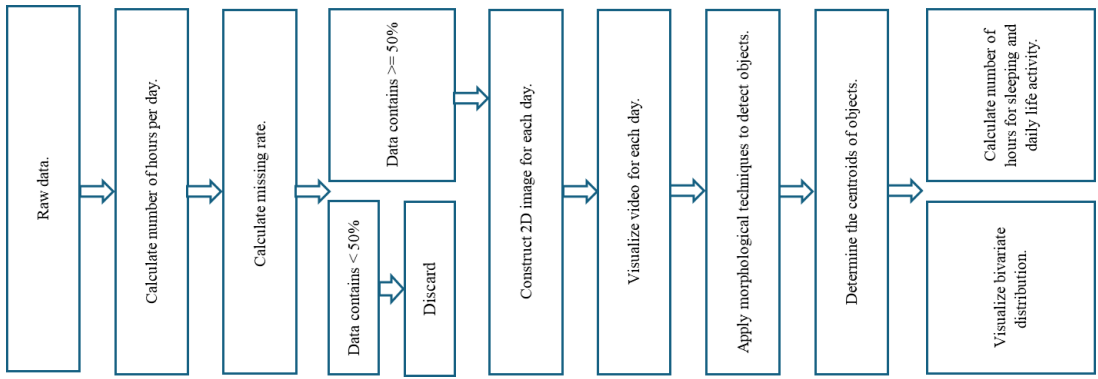

**Figure 2.** Block diagram for monitoring human behavior.

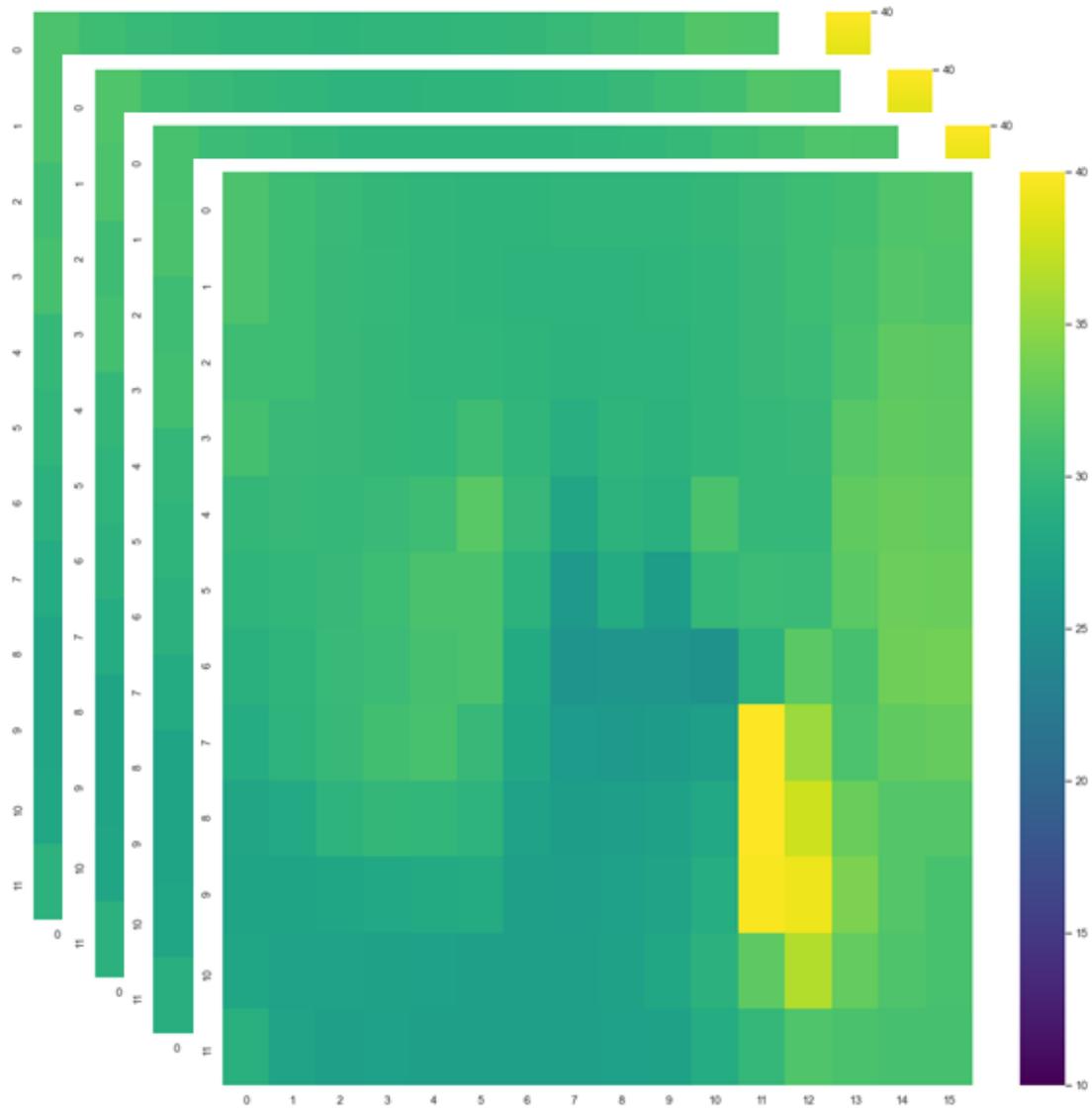

**Figure 3.** They constructed 2D images on the 10th of April.

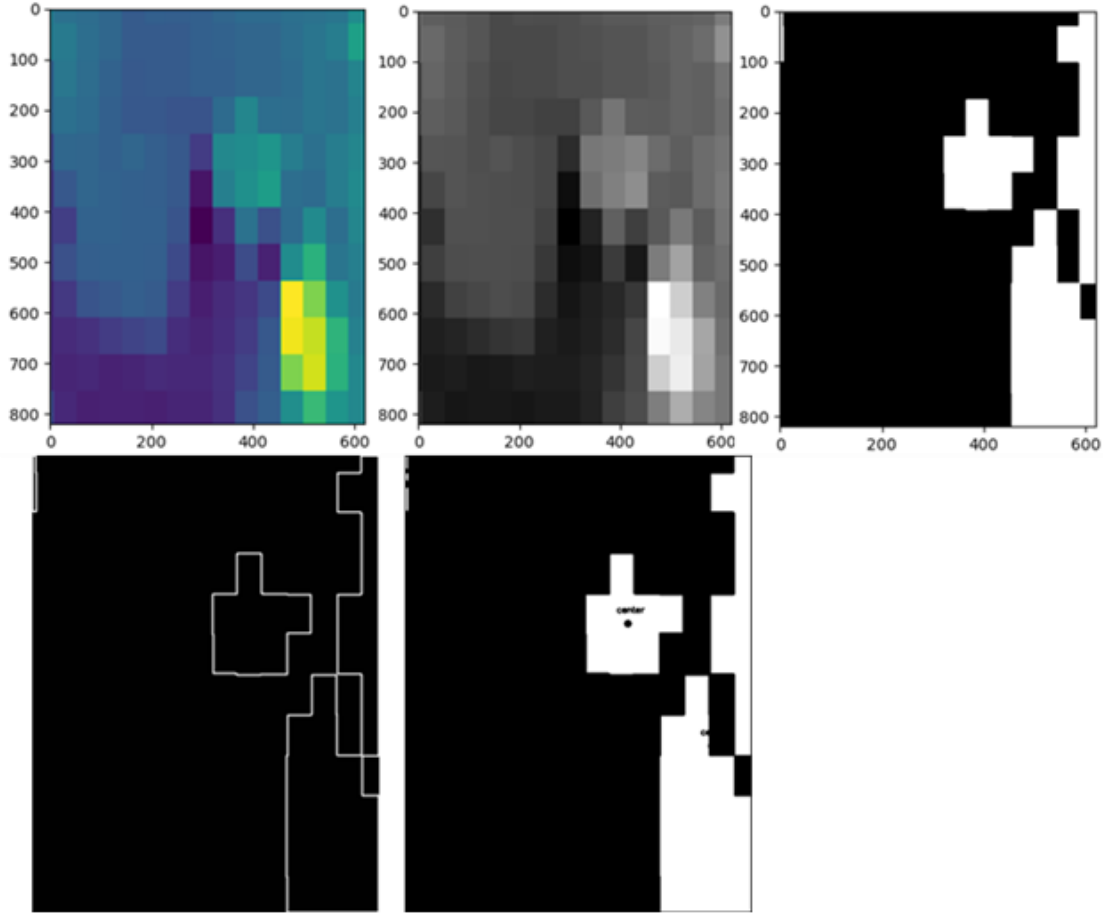

**Figure 4.** The steps to determine the centroids of objects in a 2D image.

## IV. RESULTS

The raw data obtained from the thermal sensor array MlX90640 included the time stamp and the sensor's corresponding data. The raw data was obtained for 35 days. The percentage of missing data was seven days (20%) out of the entire monitoring period of 35 days; the number of hours containing acquired data per day was plotted in Figure 5. The days with data ≥12 hours were taken into account. For each day satisfying such criteria (data ≥12 hours), multiple 2D images were constructed to form a video stack per day. As an example,

Figure 6 showed two non-interpolated (16x12) images representing two different activities performed by the person, typically the sleeping Activity and daily Activity.

The output of the morphological operations, along with the interpolated (800x600) image to find all the centroids of all objects presented in the 2D images forming the video stack for the 8th of April as an example, are presented in Figure 7. These centroids were projected on a scatter plot to represent all possible spatial locations where the person was offered throughout the day. It was worth noting that the resultant objects might describe the person himself or the heater found inside the room. To remove all the objects representing the heater, the heater's spatial location from the scatter plot was discarded, as represented by Figure 8. The centroids for the person for all days were described in Figure 9-Figure 15.

Based on the spatial locations of all the centroids in the bed area, the timestamps were concluded at which the sleeping activity was detected for the entire monitoring period in days, as shown in Figure 16. Similarly, the timestamps at which the daily activity was detected for the whole monitoring period were concluded and shown in Figure 17. The number of hours of sleeping activity and the daily per day for the person were calculated using the data obtained in the previous figures, which are shown in Figure 18 and Figure 19, respectively.

Bivariate distribution was applied to represent the spatial distribution of the activity for the person. The bivariate distribution was calculated by calculating the discrete probability for the scattered centroids and their equivalent bivariate distribution. Afterward, these bivariate distributions were plotted as heat maps to highlight the spatial locations at which there was a significant probability of finding the person in that specific spatial location. Logarithmic normalization was applied to the centroid data to adjust the scale of the resulting spatial distribution as an image. Such a step of logarithmic normalization led to the advantage of standardizing the images and the feasibility of comparing them. Following, the bivariate distribution results per day were presented in detail.

As shown in Figure 20, there were two bright spots, meaning the person had carried out activities in these spatial locations. The higher brightness of the right-hand side spot compared to the lower brightness of the left-hand side recommended that the person do more activity at that spatial location. This conclusion was reinforced by the higher horizontal probability distribution value for the right-hand side peak than the corresponding left-hand side peak.

As shown in Figure 21, there were two bright spots on the 7$^{th}$ of April, which meant that the person had carried out activities in two spaces in the room, as the person was active in these spaces and the rest of the room was empty and the higher brightness of the right-hand side spot compared to less brightness of the left-hand side one recommended that the person did more activity at that spatial location. The same two bright spots were found on the 8th of April in the same spaces on the 7th of April, but on the 8th of April, small spots were observed, which meant that the person had carried out activities in new spaces. On the 9th of April, two bright spots were found and several faint spots, which meant that the person did behave in multiple spaces; the bright spot on the right was more significant than the one on the left, and two bright spots were observed and one faint spot on the 10th of April and the higher brightness of the right-hand side spot compared to less brightness of the left hand sided one recommended that the person did more activity at that spatial location.

As shown in Figure 22, there were two bright spots and three small faint spots on the 11th of April; the higher brightness of the right-hand side spot compared to less brightness of the left-hand sided one recommended that the person did more activity at that spatial location and the two bright spots were close to each other. On the 12th of April, the two bright spots were much closer together, and there were several faint spots, but the two bright spots were almost equal. There were two bright spots, and several small faint spots were found on the 13th of April; the two bright spots were connected. On the 14th of April, there were only two bright spots, and the bright spot on the right was more significant than the one on the left, which meant that the person had carried out activities in only two spaces in the room.

As shown in Figure 23, there were two bright spots and small faint spots on the 15th of April, and the bright spot on the right was more significant than the one on the left, which meant that the person had carried out a few activities in only two spaces in the room. On April 16$^{th}$, three prominent spots and three smaller dim ones appeared. The central bright spot held more significance than those on the right and left, indicating heightened activity. The 17th of April revealed numerous bright spots, suggesting activity across multiple areas in the room. By the 19th of April, two bright spots and several faint ones were observed, with the higher brightness of the spot on the right suggesting increased activity in that specific location compared to the dimmer spot on the left.

As shown in Figure 24, there were many bright spots in different spaces on the 20$^{th}$ of April, which meant that the person had carried out new and more behaviors in many spaces in the room. There was one bright spot and many faint spots on the 21$^{st}$ of April, and there was only one bright spot and a few small faint spots on the 22$^{nd}$ of April. Finally, there were two bright spots and three faint spots on the 23$^{rd}$ of April, and the bright spot below was more significant than the one above.

As shown in Figure 25, there were two bright spots and other faint spots on the 24$^{th}$ and 25$^{th}$ of April. The higher brightness of the right-hand side spot compared to the lower brightness of the left-hand side recommended that the person had carried out more activity at that spatial location for two days. On the 26th of April, there were two bright spots and another two faint spots, and the bright spot below was more significant than the one above. There was only one bright spot, and several faint spots were found on the 27th of April, which meant that the person had carried out activity in new and different spaces.

As shown in Figure 26, on the 28th of April, there were several bright spots and faint spots, and the activity was concentrated in several spaces of the room; on the 29th of April, there were one bright spot and three faint spots. One big bright spot and other faint spots were observed on the 30th of April. One bright spot and several small faint spots were observed on the 1st of May.

As shown in Figure 27, the activity of the person was concentrated in three spaces on the 2nd of May, and the highest activity was concentrated on the right on the 3rd of May; there were several bright spots and faint spots, and the highest activity was focused on the right. There was one bright spot and multiple faint spots on the 5th of May, and the person's activity was concentrated on the right. On the 6th of May, there were several bright spots and fait spots with the higher brightness of the right-hand side spot compared to the less brightness of the left-hand side; one recommended that the person did more activity at that spatial location.

The spatial distribution of the person's activity for all days is represented in Figure 21-Figure 27. For sleeping Activity and daily Activity, the comparison between our results and the study of H. M. Ahmed et al. [43] was presented in Table 1. For sleeping and daily activities, the comparison between our results and the study of H. M. Ahmed et al. was presented in Figure 28.

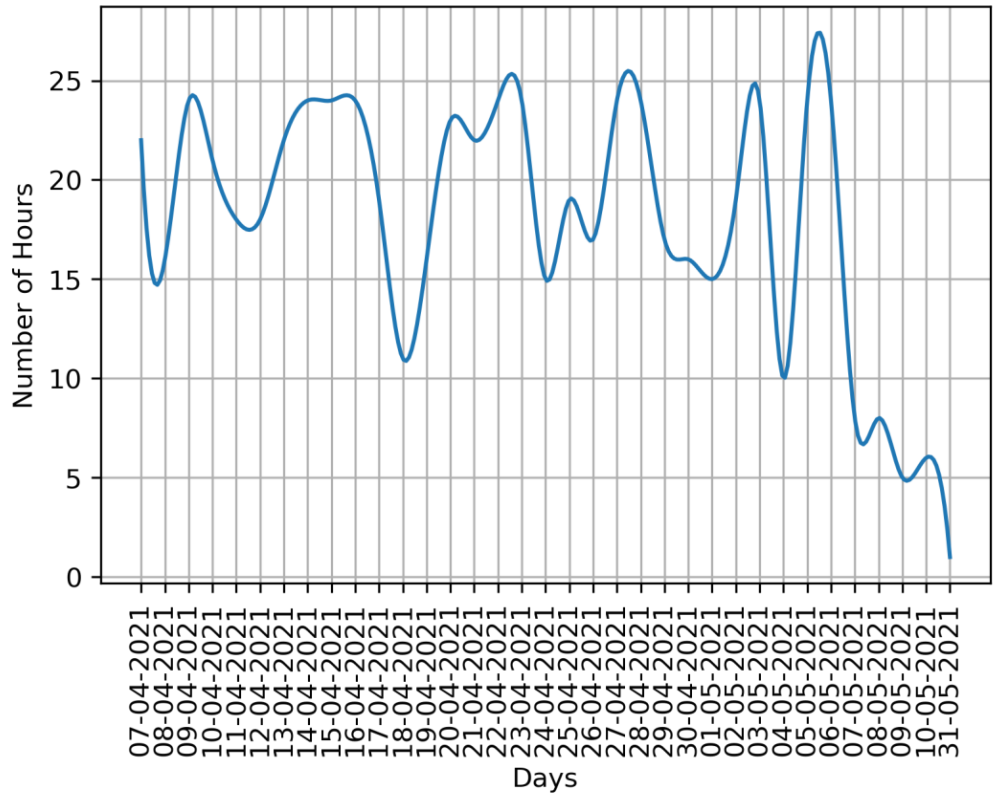

**Figure 5.** Number of hours containing acquired data per day.

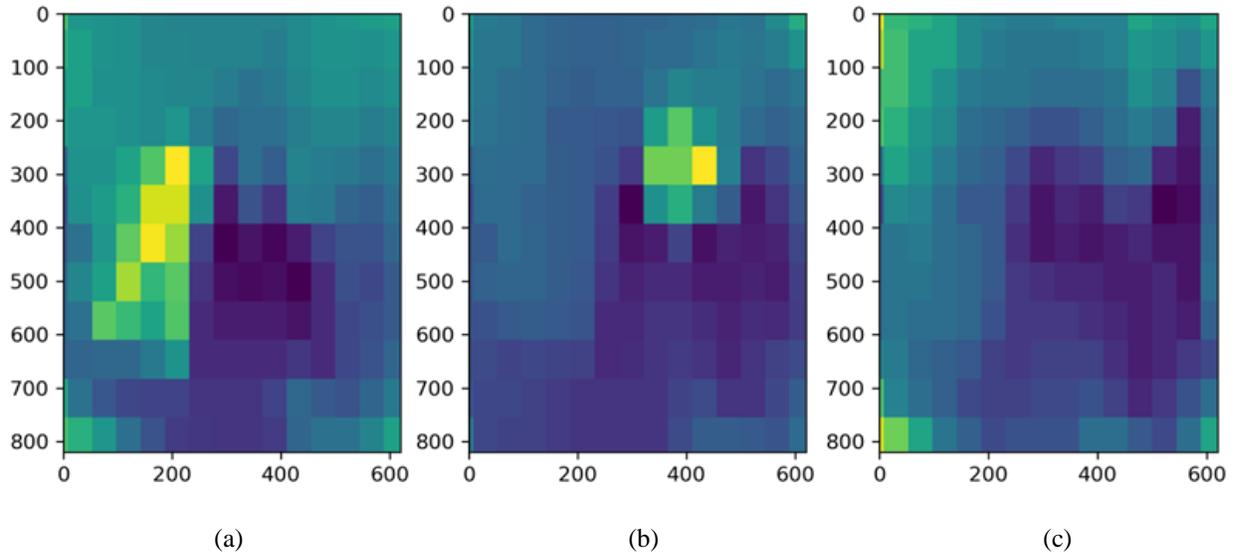

Figure 6. (a) Constructed 2D Image showing Sleeping activity, (b) Constructed 2D Image showing Daily Activity, and (c) Constructed 2D image showing No Activity [41].

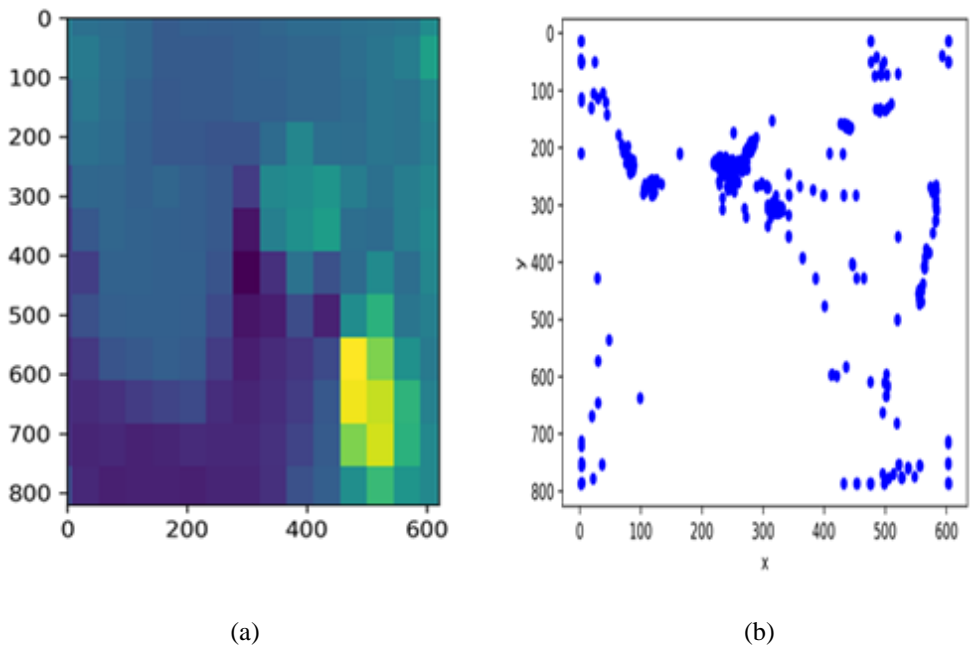

**Figure 7.** (a) Constructed 2D Image, and (b) The centroids of all objects (person, heater) which are existing in 2D images during the 8th of April.

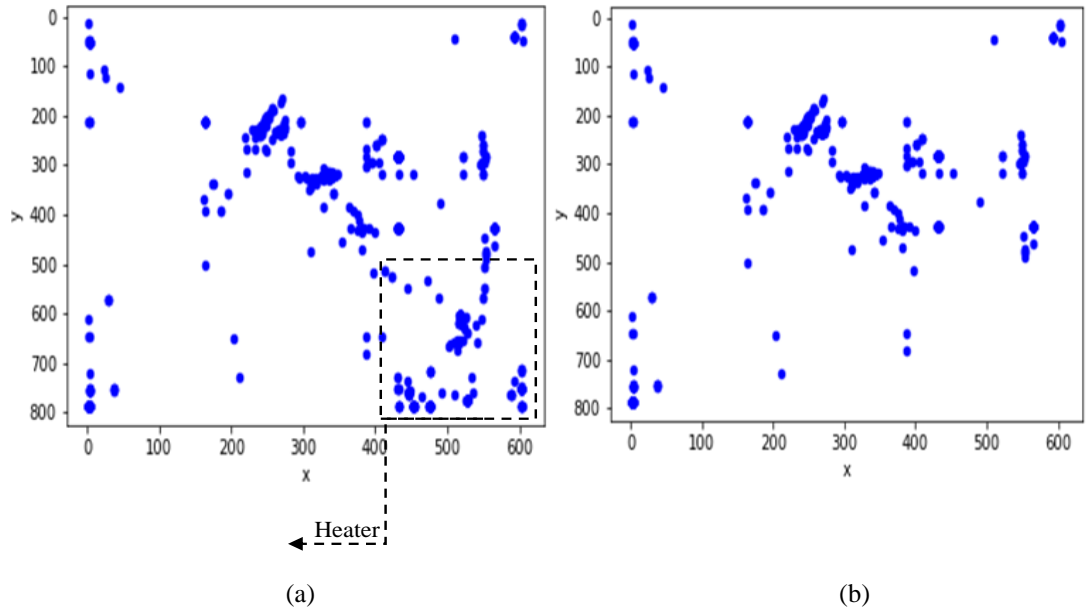

**Figure 8.** (a) The centroids of all objects (person, heater) during the 7th of April, and (b) the centroids of the person only during the 7th of April.

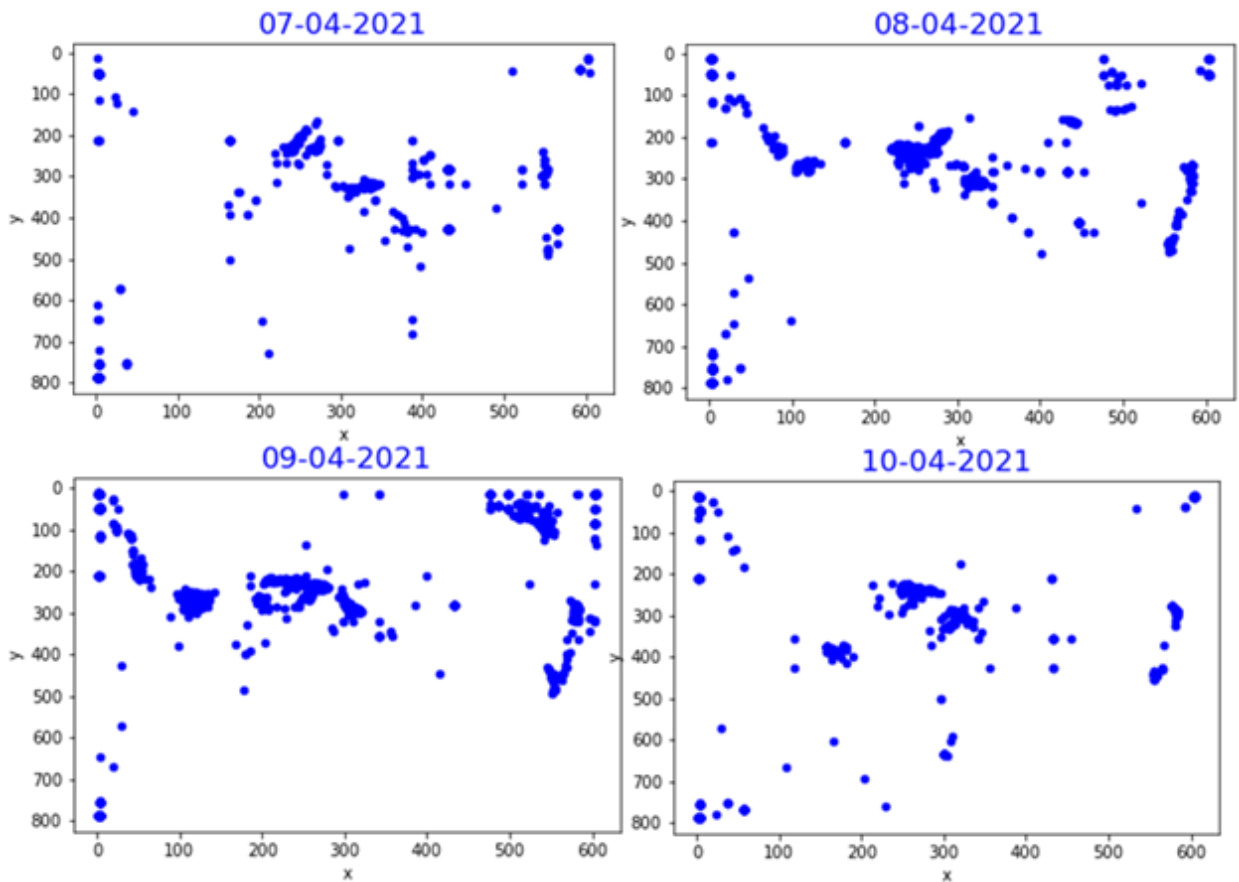

**Figure 9.** The centroids of person for 7th, 8th, 9th, and 10th April.

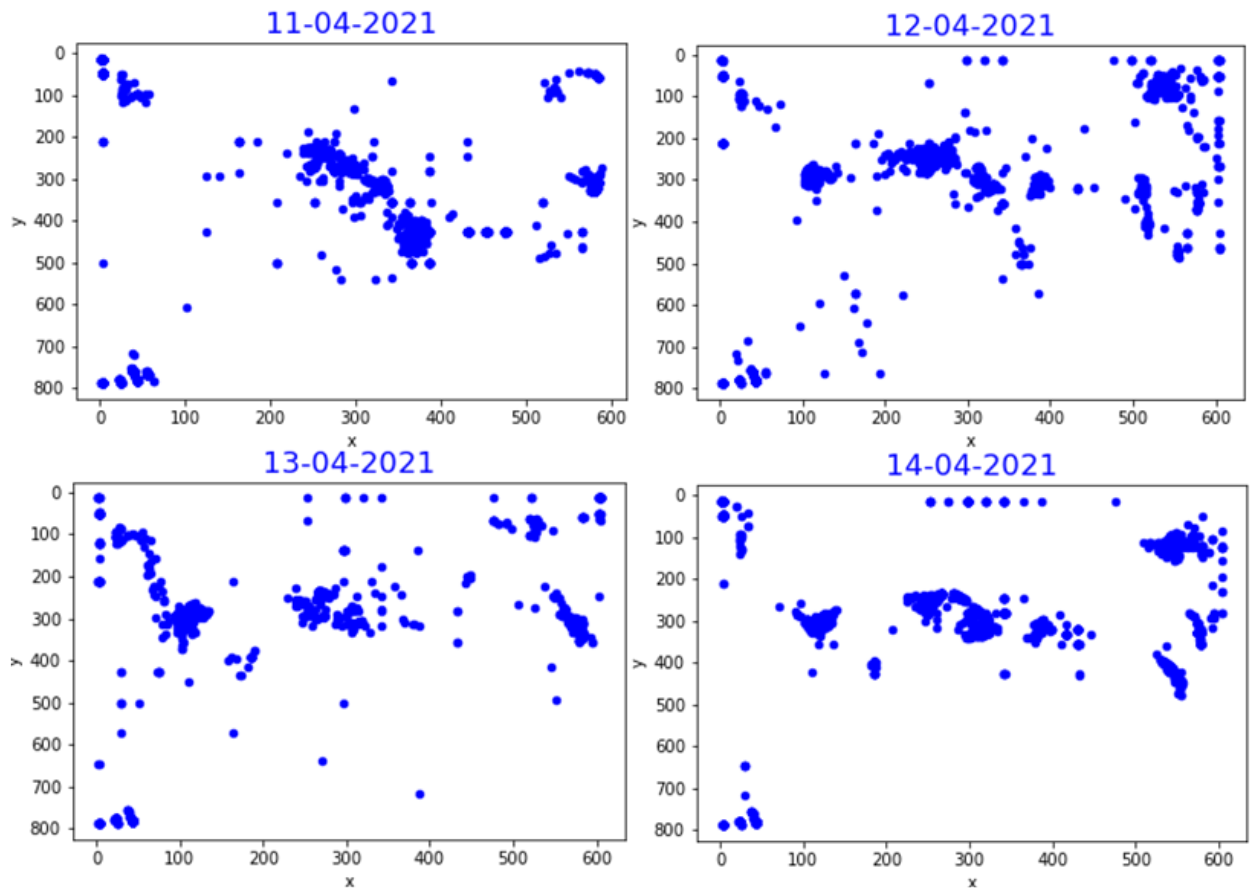

**Figure 10.** The centroids of person for 11[th], 12[th], 13[th], and 14[th] April.

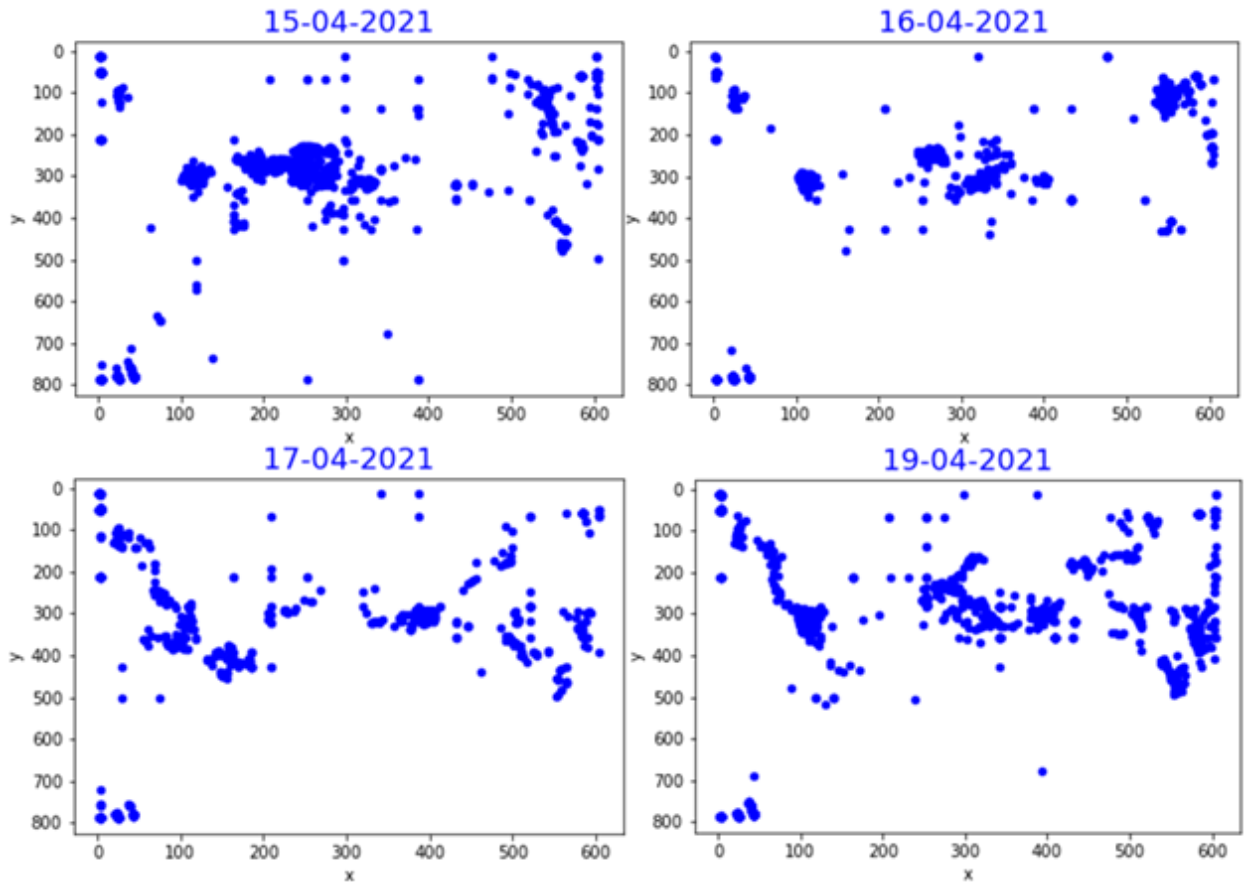

**Figure 11.** The centroids of person for 15[th], 16[th], 17[th], and 19[th] April.

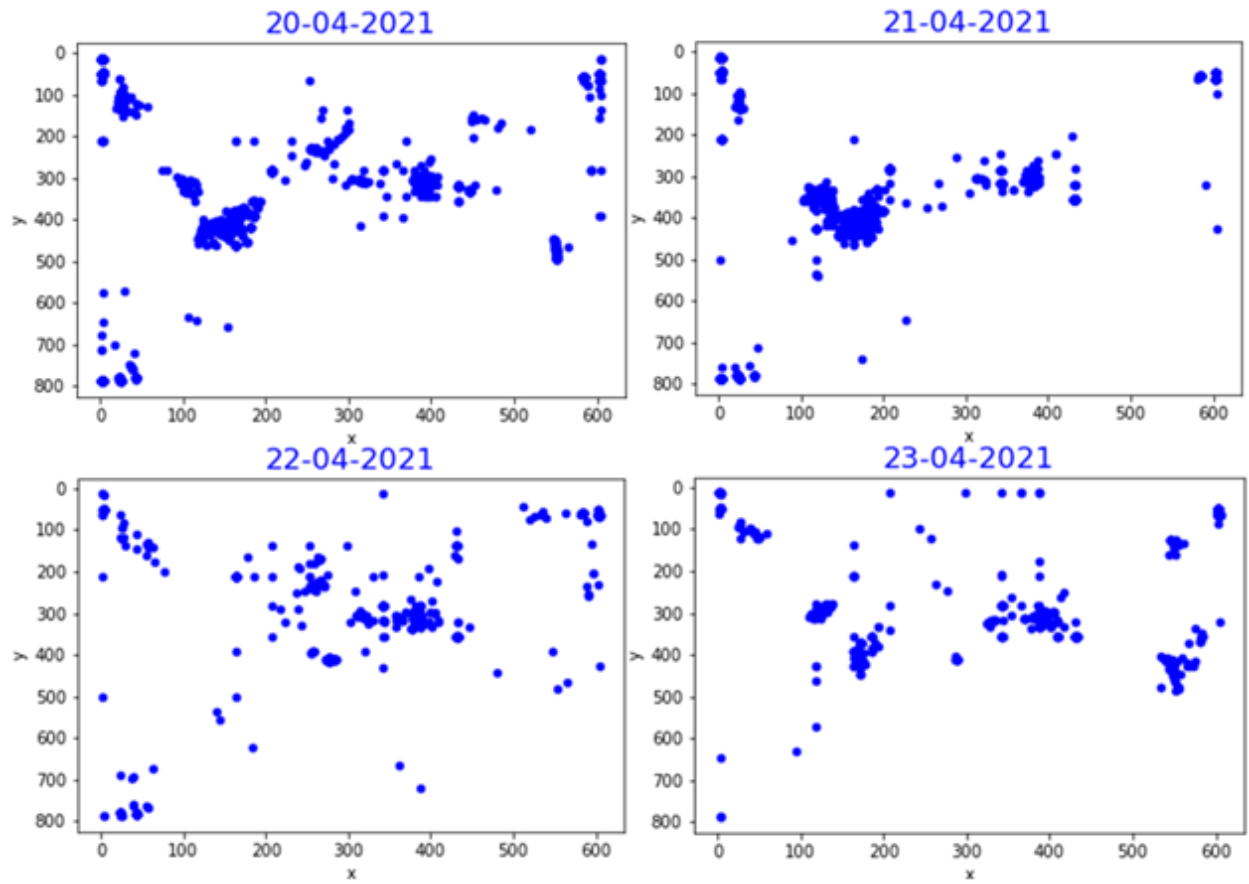

**Figure 12.** The centroids of person for 20th, 21st, 22nd, and 23rd April.

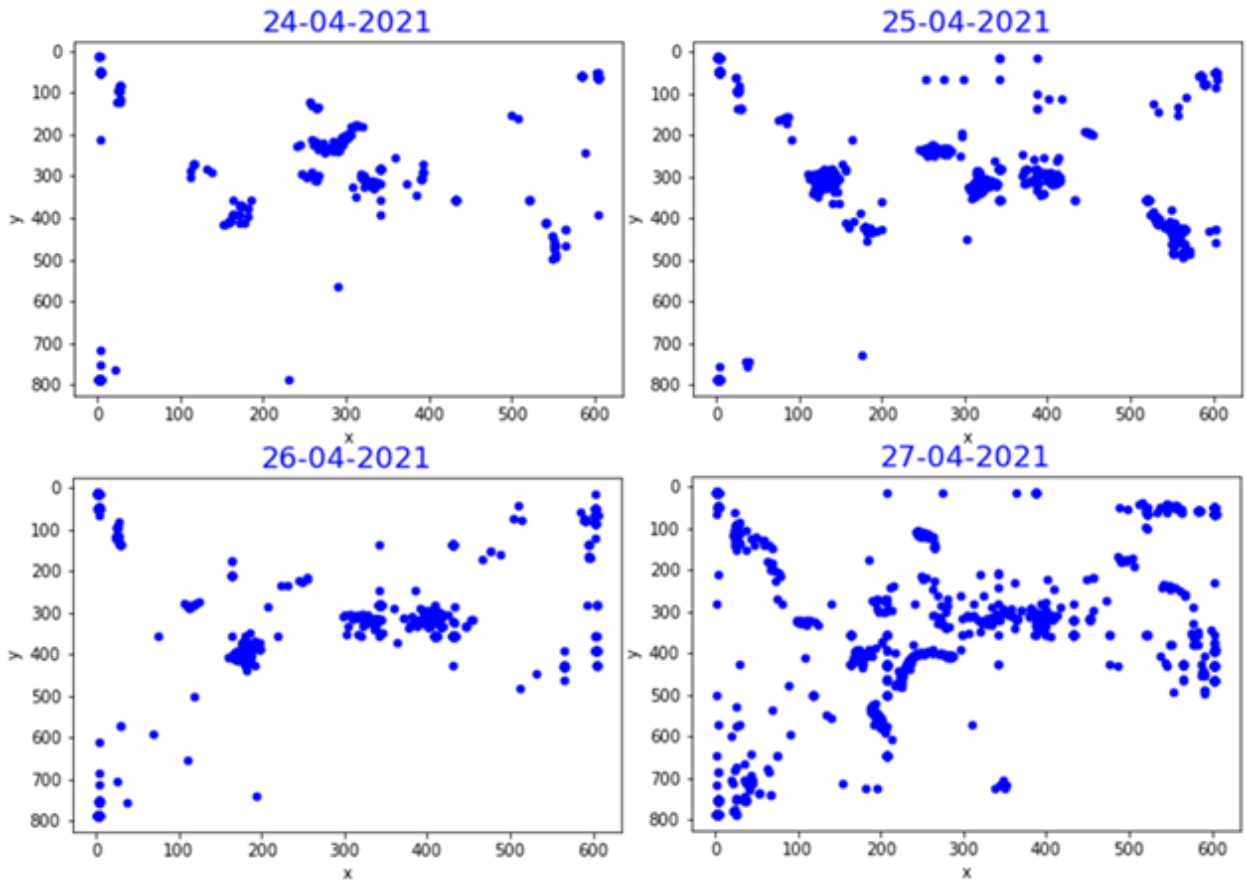

**Figure 13.** The centroids of person for 24th, 25th, 26th, and 27th April.

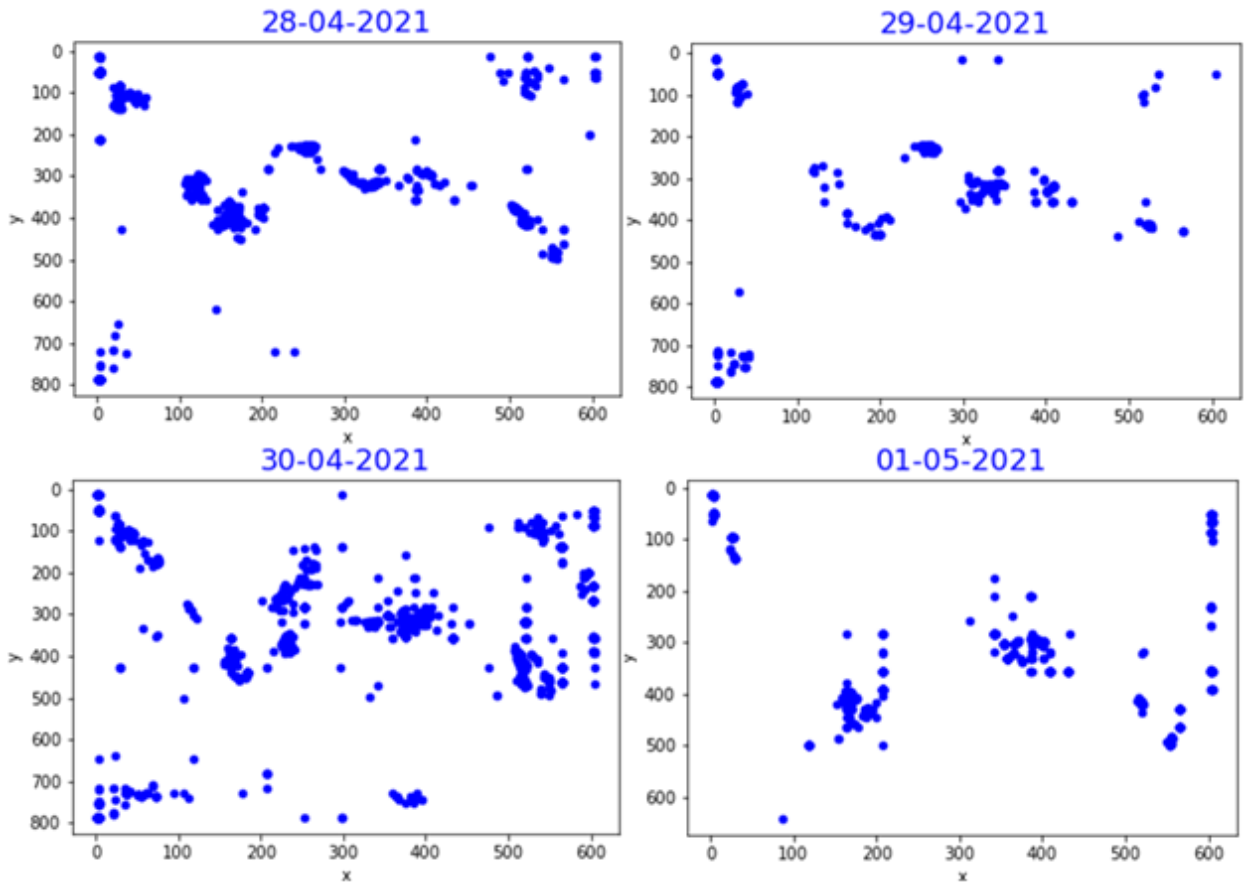

**Figure 14.** The centroids of person for 28th, 29th, 30th April, and 1st May.

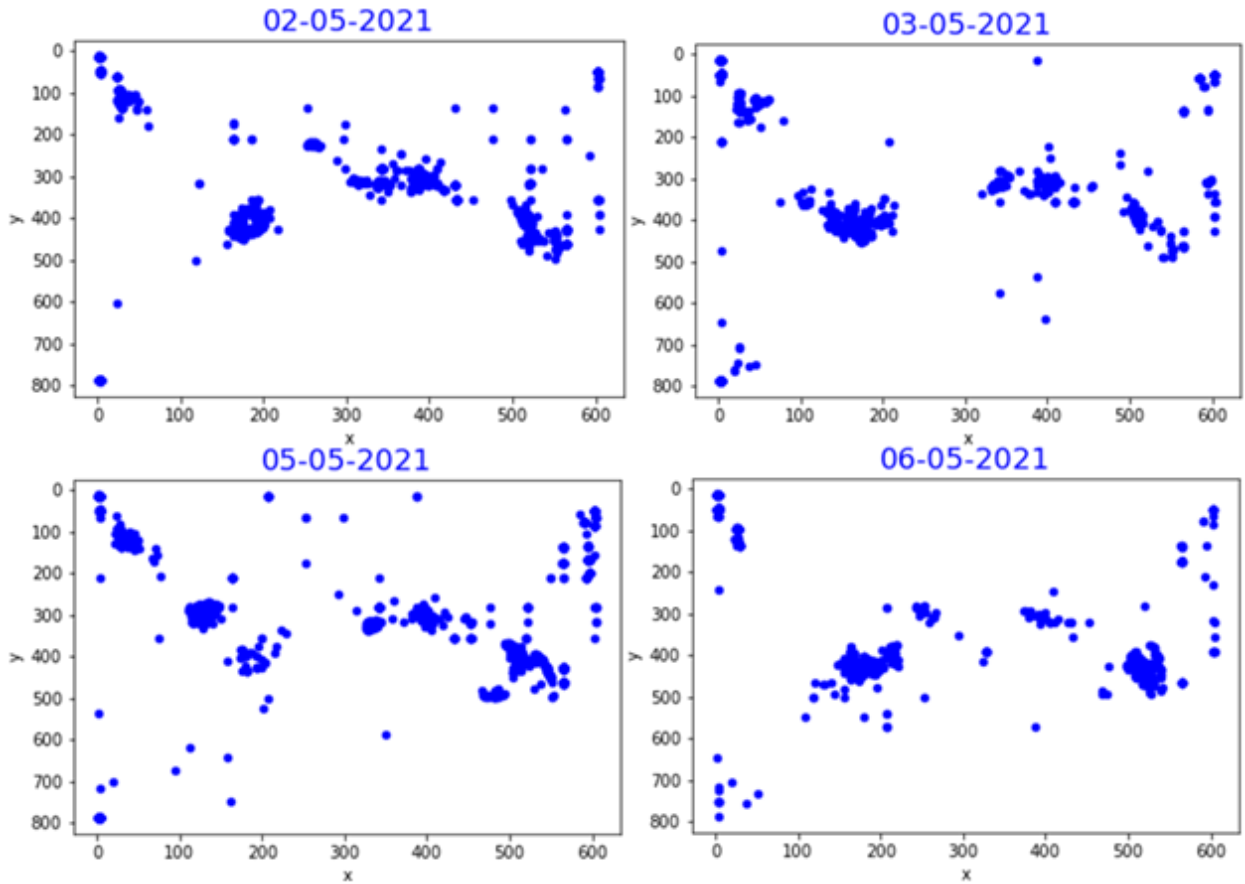

**Figure 15.** The centroids of person for 2<sup>nd</sup>, 3<sup>rd</sup>, 5<sup>th</sup>, and 6<sup>th</sup> May.

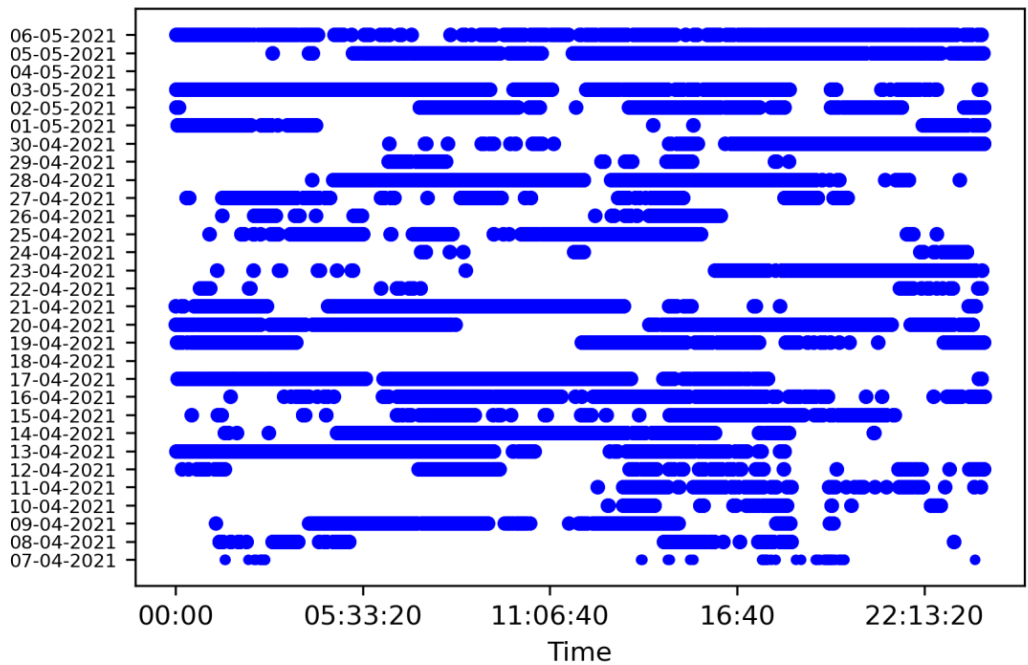

**Figure 16.** The timestamp of sleeping activity for all days.

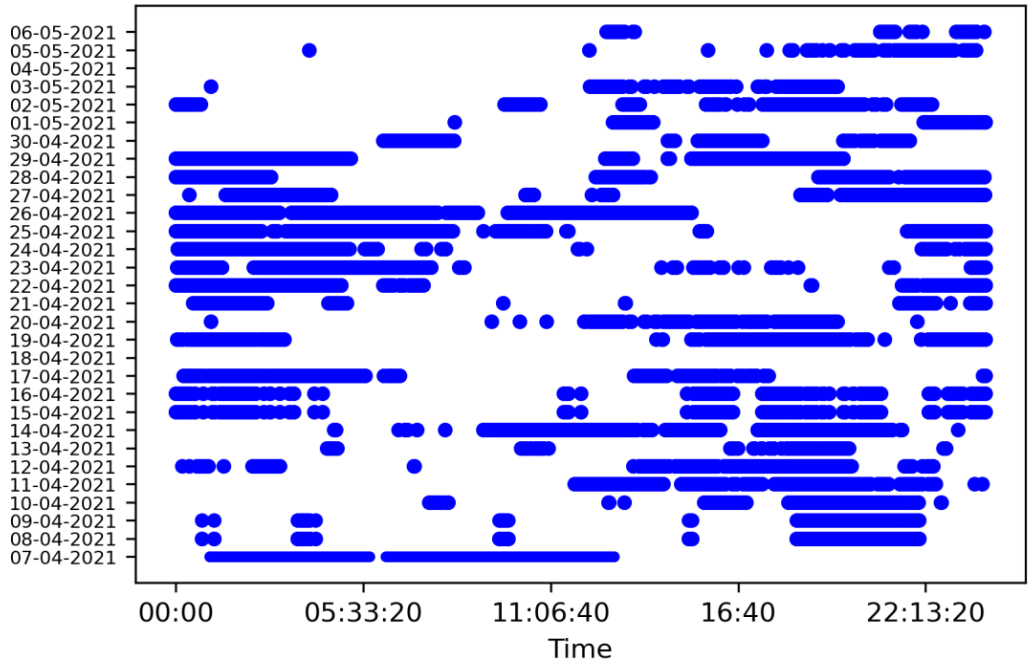

**Figure 17.** The timestamp of daily activity for all days.

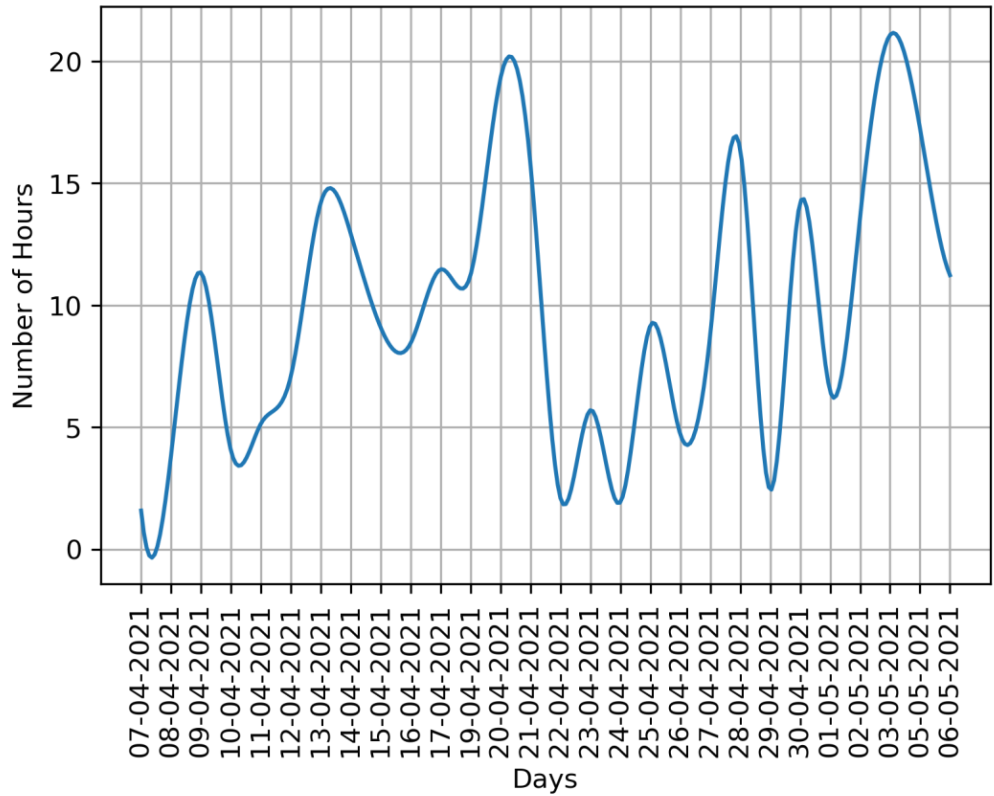

**Figure 18.** The number of hours for sleeping activity per day for the entire monitoring period of 28 days.

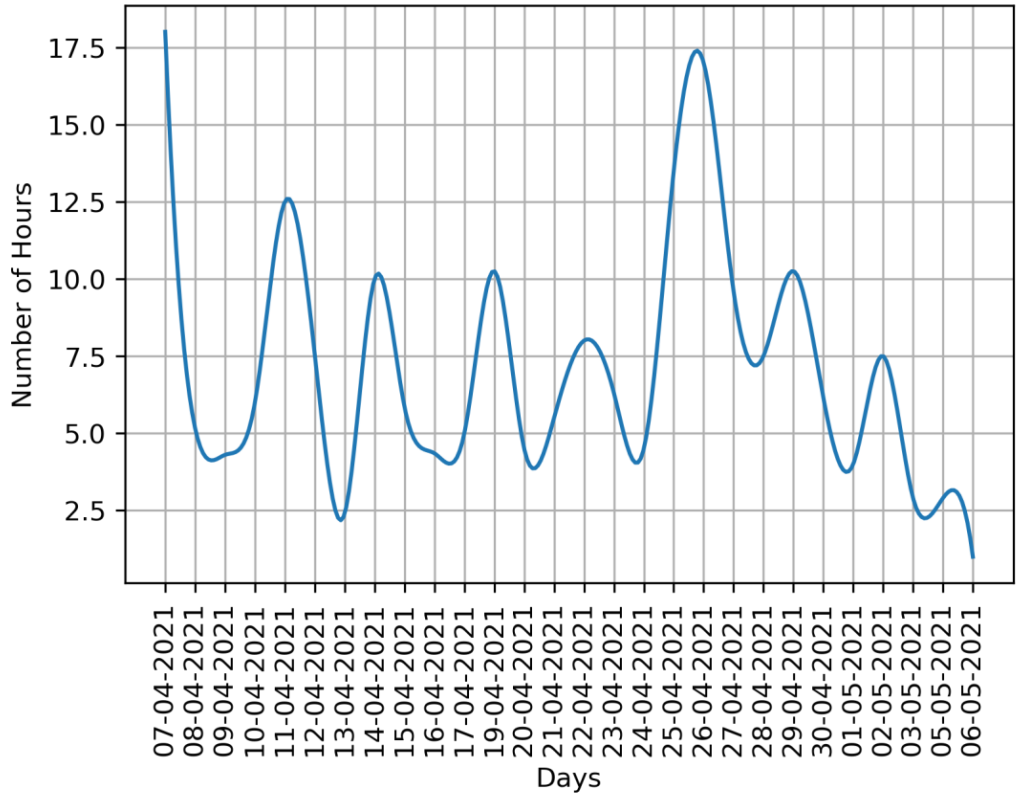

**Figure 19.** The number of hours for daily activity per day for the entire monitoring period of 28 days.

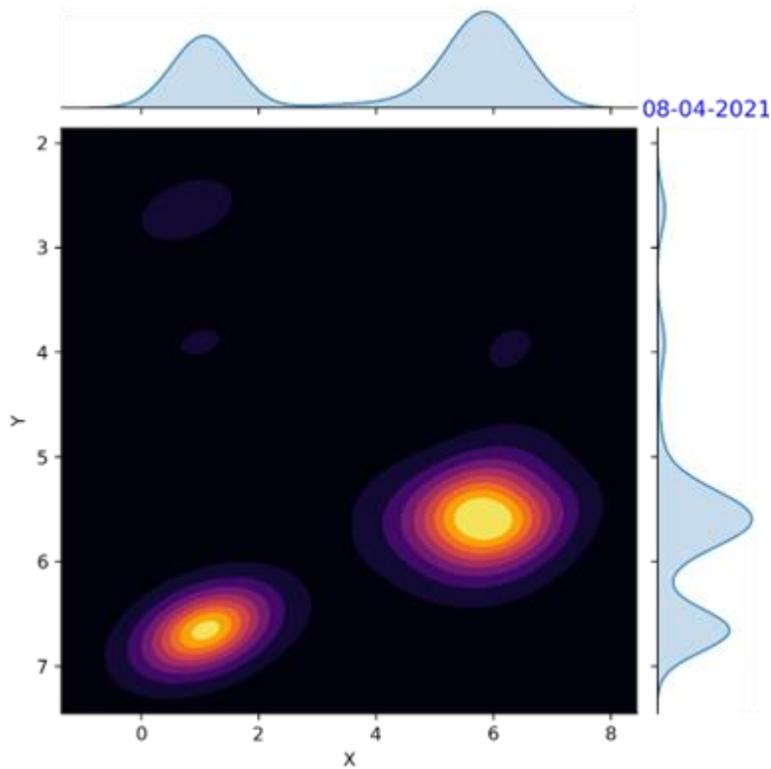

**Figure 20.** The bivariate distribution during the 8th of April.

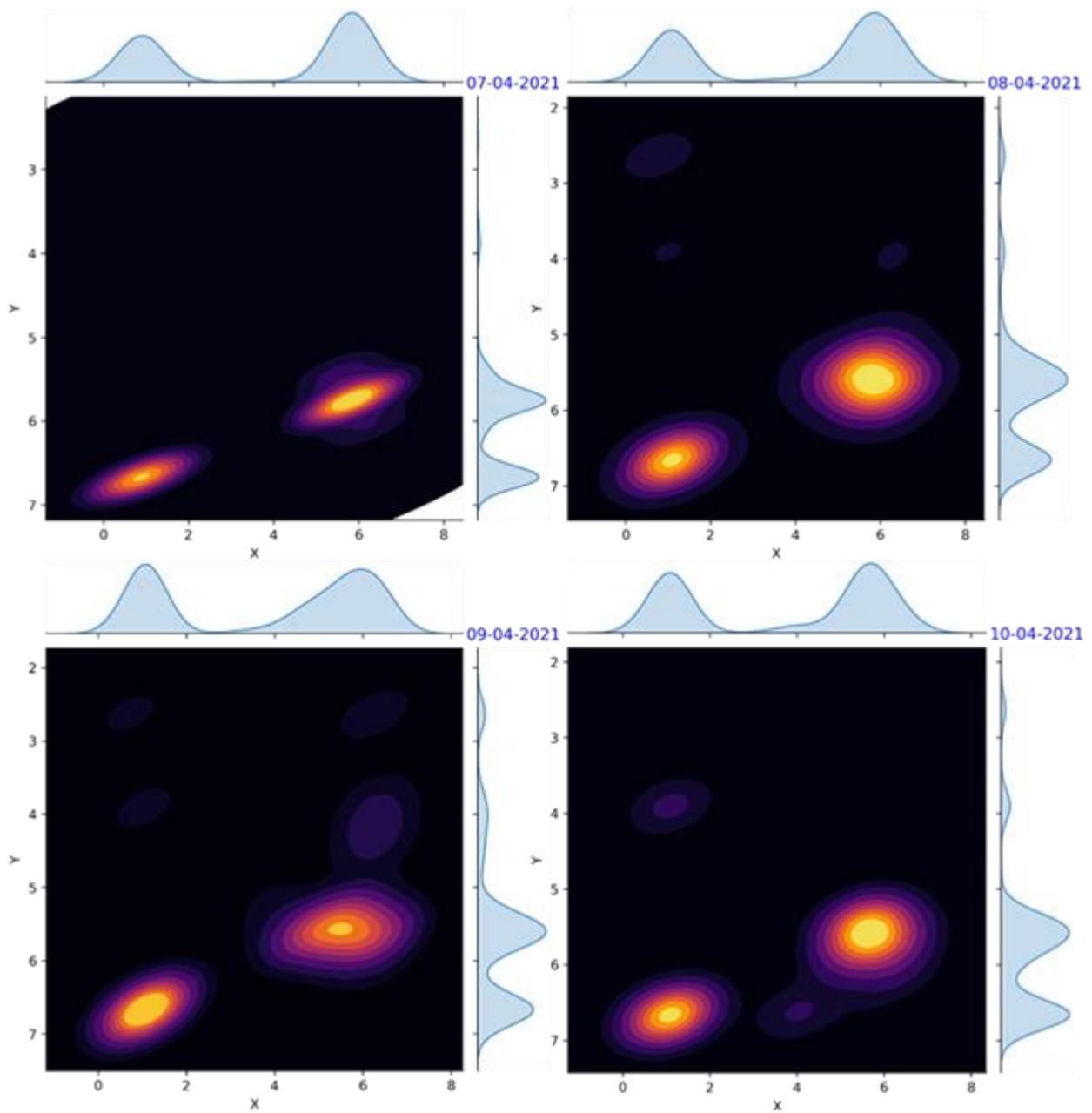

**Figure 21.** The bivariate distribution for 7th, 8th, 9th, and 10th April.

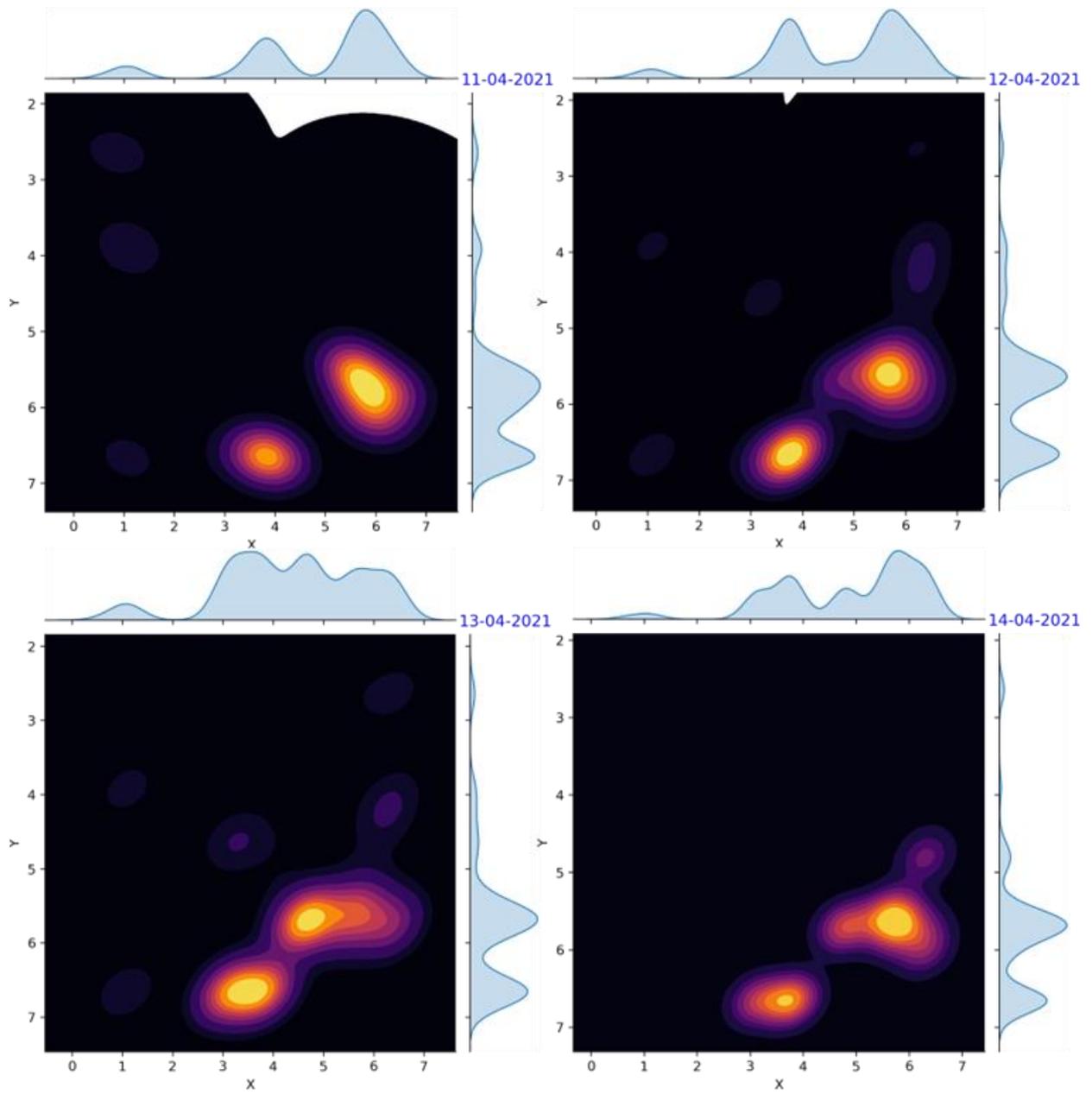

**Figure 22.** The bivariate distribution for 11[th], 12[th], 13[th], and 14[th] April.

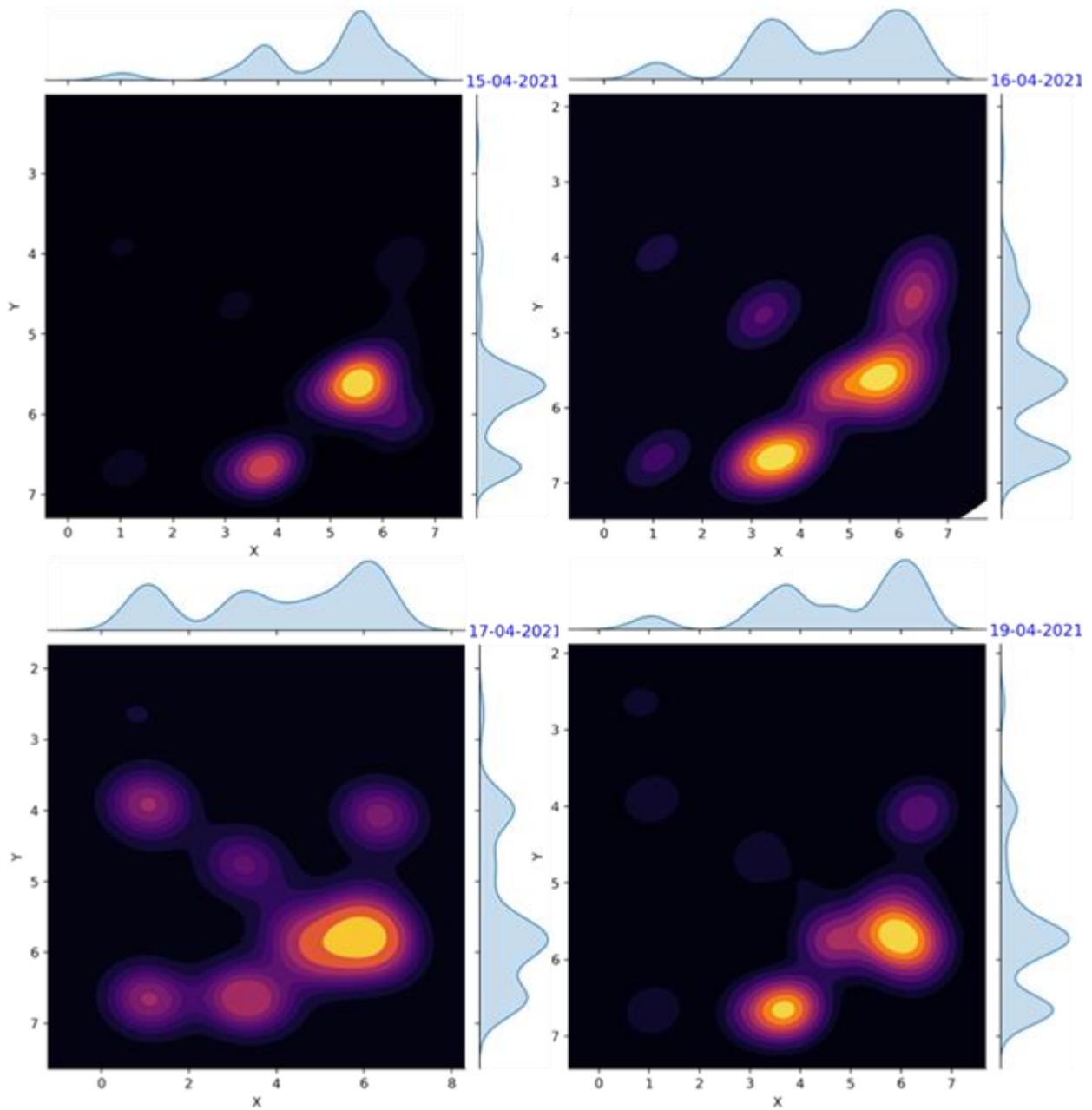

**Figure 23.** The bivariate distribution for 15[th], 16[th], 17[th], and 19[th] April.

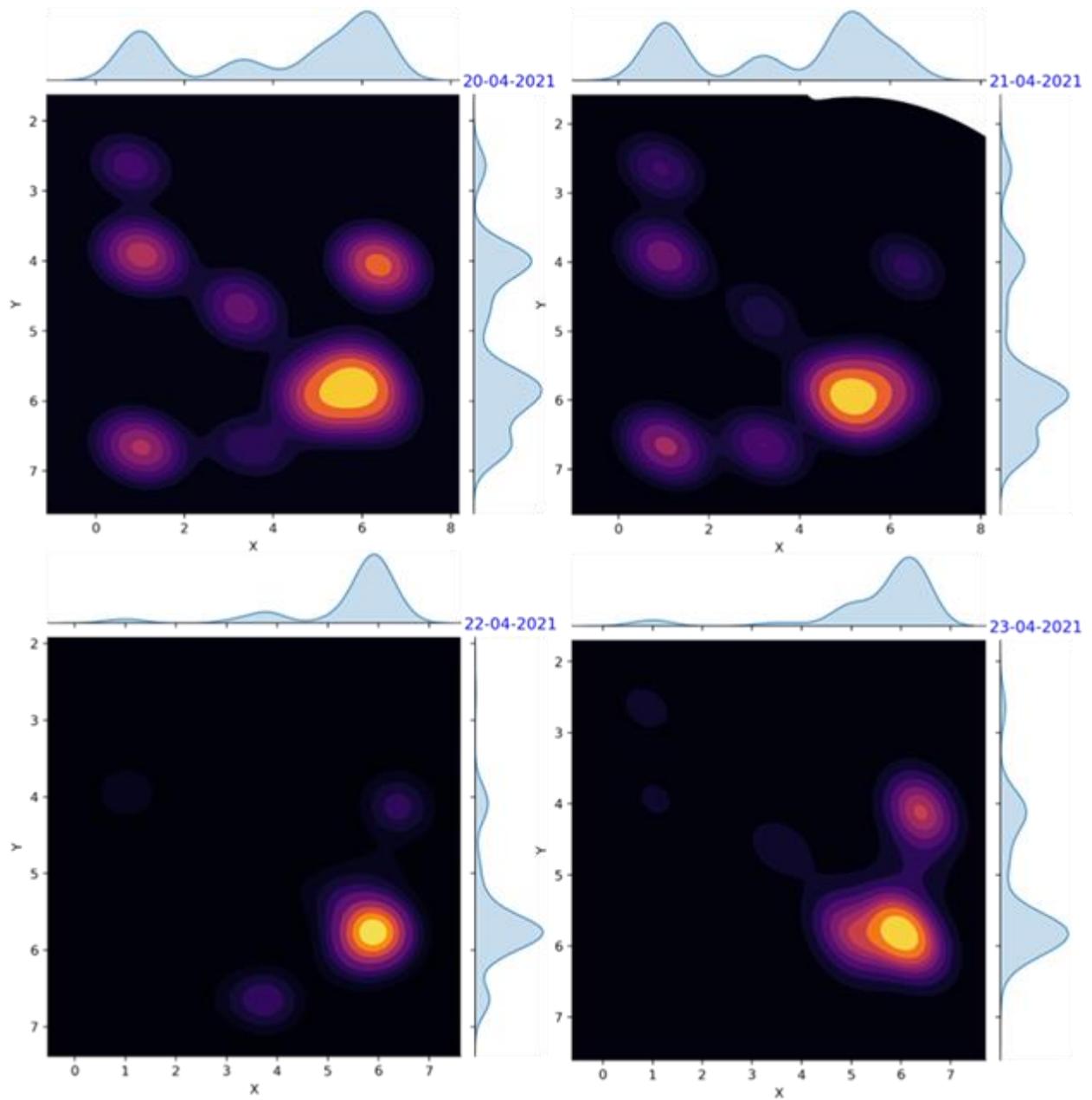

**Figure 24.** The bivariate distribution for 20[th], 21[st], 22[nd], and 23[rd] April.

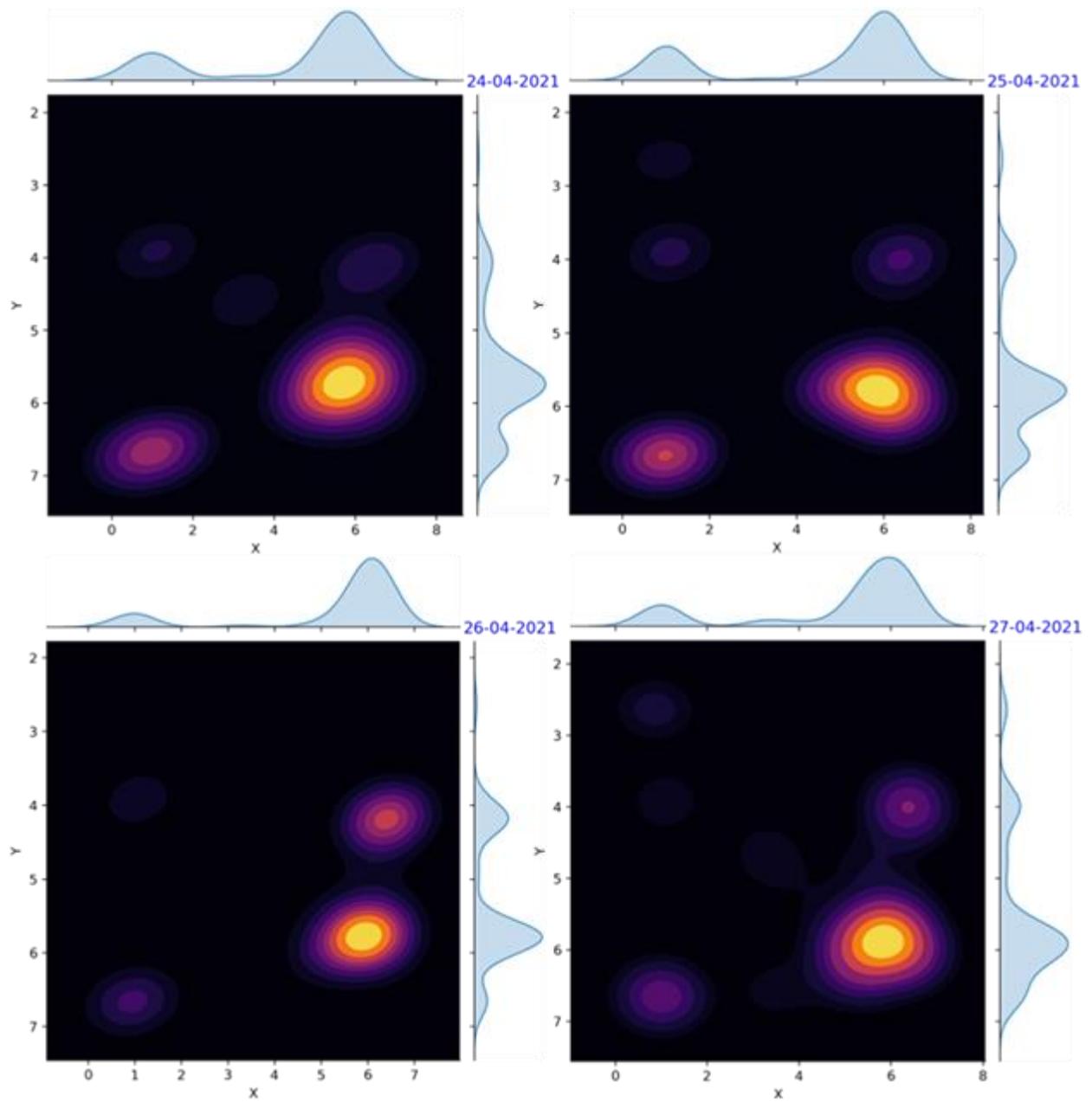

**Figure 25.** The bivariate distribution for 24[th], 25[th], 26[th], and 27[th] April.

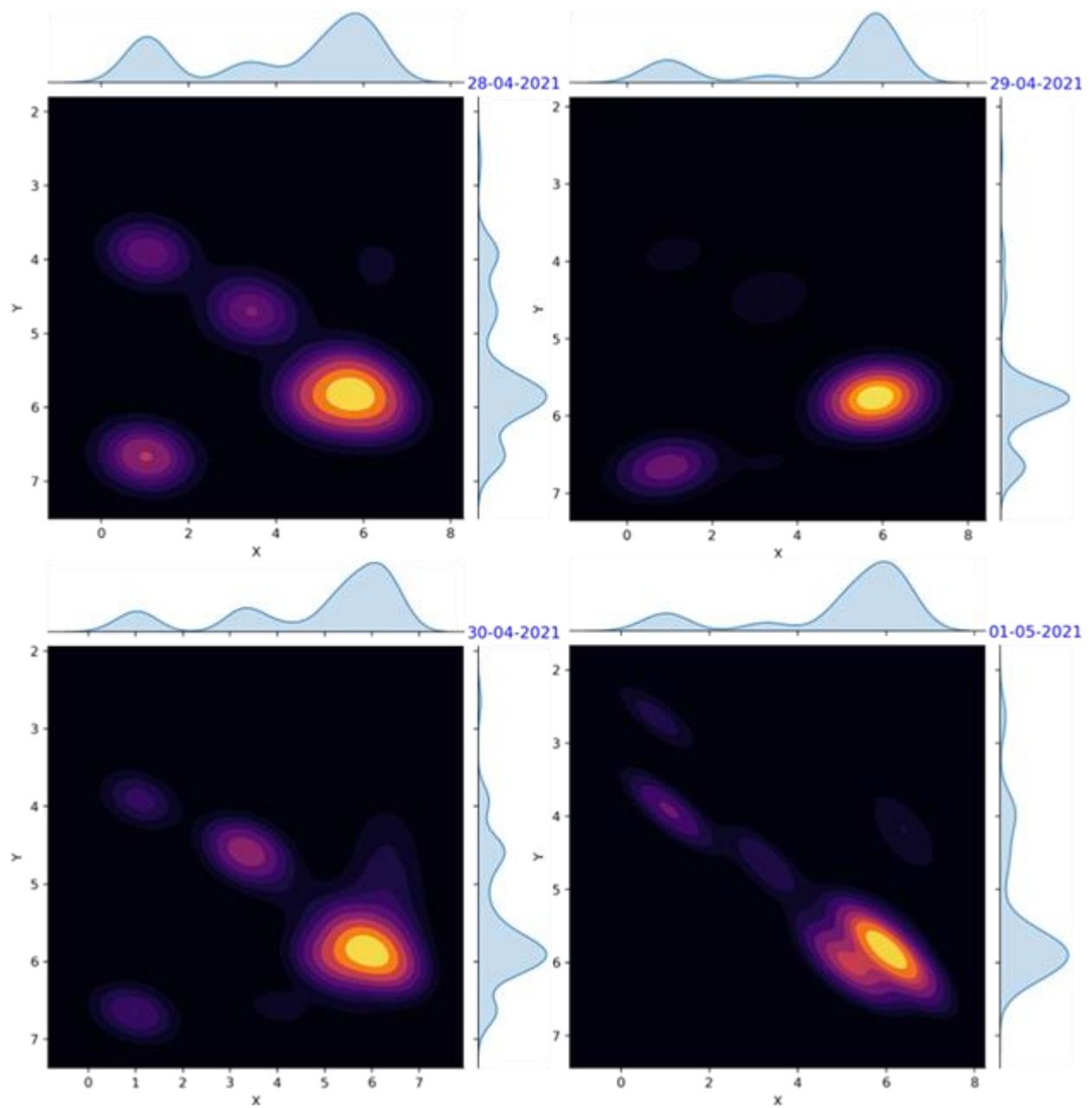

**Figure 26.** The bivariate distribution for the 28th, 29th, 30th of April, and 1st of May.

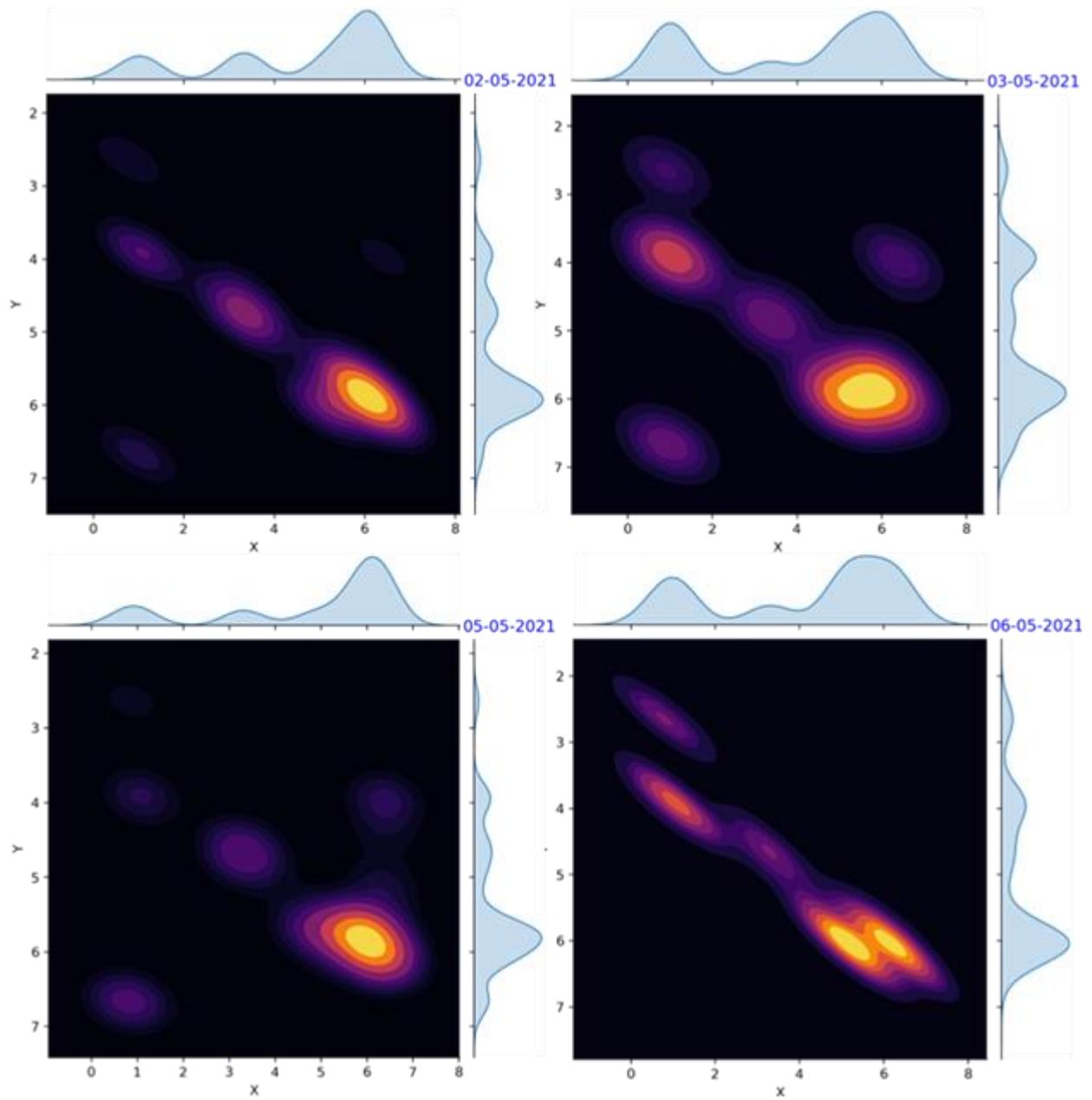

**Figure 27.** The bivariate distribution for 2nd, 3rd, 5th, and 6th May.

**Table 1.** The number of hours for sleeping and daily activities for 10 days for our study and H. M. Ahmed *et al.* [43].

|  |  | Sleeping Activity |  | Daily Activity |  |
|---|---|---|---|---|---|
|  |  | **Our study** | **H. M. Ahmed et al. [43]** | **Our study** | **H. M. Ahmed et al. [43]** |
| Monitoring Day | 7th | 1.58 | 11.5 | 18 | 11.5 |
|  | 8th | 3.85 | 2 | 5.16 | 6.5 |
|  | 9th | 11.33 | 10 | 4.3 | 4 |
|  | 10th | 4.03 | 12 | 6 | 5 |
|  | 11th | 5.15 | 5 | 12.5 | 6 |
|  | 12th | 7.1 | 7.5 | 7.55 | 6 |
|  | 13th | 14.22 | 10 | 2.4 | 4 |
|  | 14th | 12.92 | 13 | 10 | 10.5 |
|  | 15th | 9.08 | 9.5 | 5.85 | 7 |
|  | 16th | 8.45 | 11.5 | 4.35 | 8 |
|  | 17th | 11.47 | 8 | 5 | 9.5 |

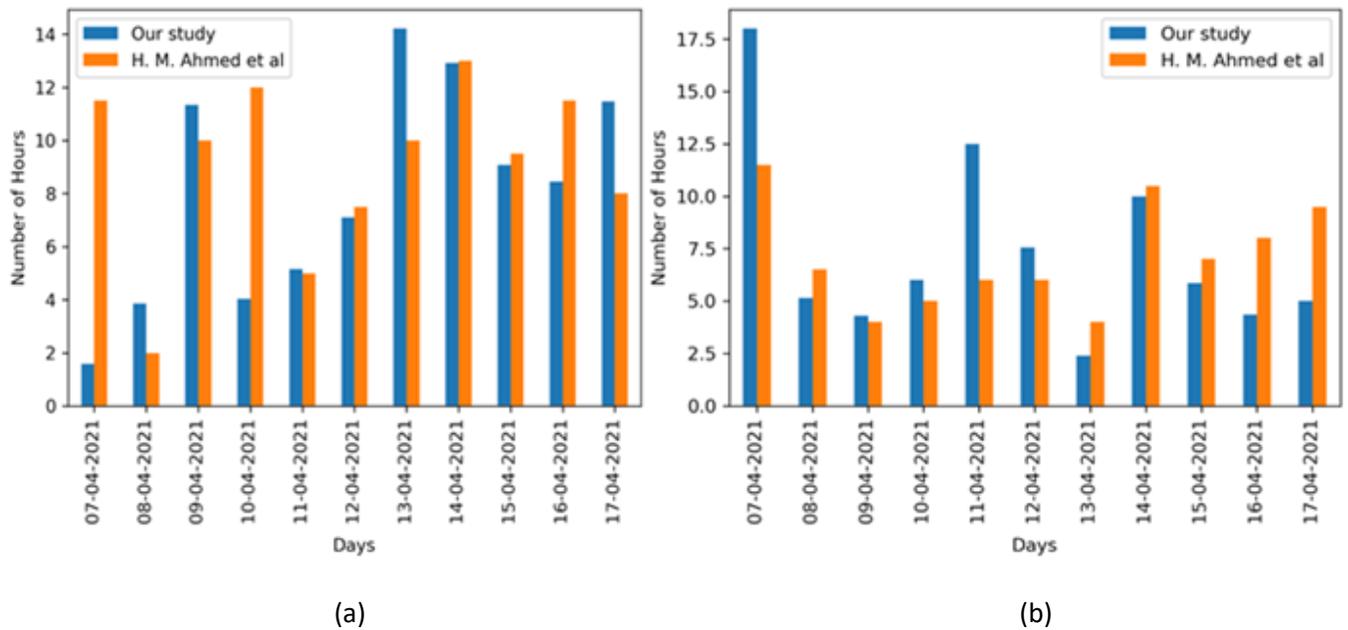

(a)          (b)

**Figure 28.** (a) The number of hours for sleeping activity of our study and H. M. Ahmed *et al.*, and (b) The number of hours daily activity of our research and H. M. Ahmed *et al*.

### V. DISCUSSION

The raw data was obtained for 35 days. The percentage of missing data for each day was determined by calculating the number of days containing data of more than or equal to 12 hours for each day and calculating the number of days containing data of less than 12 hours. These calculations showed that the number of days that included more than or equal to 12 hours of data was 28 days, and the days that contained less than 12 hours of data were seven days. The percentage of missing data for each day was 20%.

The raw data from the sensor in 1D, 1D array was converted to a 2D array to construct a 2D image for each day from greater than or equal to 50% data days and visualized each day as a video. The frames from the video were extracted for each day, and morphological operations were applied to calculate the centroids for all objects in 2D video frames. The centroids of all objects were calculated to track the person's behavior.

The timestamp of sleeping activity was shown in Figure 16, which showed that the sleep activity of the person during the $7^{th}$, $8^{th}$, $10^{th}$, $22^{nd}$, $24^{th}$, $26^{th}$, $29^{th}$, and $30^{th}$ of April interrupted sleep abnormal activity, but during other days, the sleeping activity was continuous and normal. The timestamp of daily activity, as shown in Figure 17, showed that the daily activity of the person during the $8^{th}$, $9^{th}$, $13^{th}$, $1^{st}$, $2^{nd}$, $3^{rd}$, $5^{th}$, and $6^{th}$ was decreased and abnormal, which meant that the person was not spending his time in the space or was sleep. However, on other days, the daily activity was regular, so the representation of timestamps was essential to track the sleeping and daily activities and classify if this activity was normal or abnormal. The box plot of sleeping activity was calculated as presented in Figure 29, which shows the average sleeping activity is 9.6 hours, and the number of daily activity was calculated as presented in Figure 29, which showed the average daily activity was 7.5 hours. The sleeping and daily activities were determined by whether normal or abnormal.

As already mentioned, the bivariate distribution showed the relation between the behavior or activity of the person and the spatial location. As shown in Figure 21, the $7^{th}$, $8^{th}$, $9^{th}$, and $10^{th}$ of April were very close to each other in behavior because each day contained two bright spots in the same proportion ,and there were also several faint points, which indicated that there were other activities carried out by the person in different spaces but in a small percentage. The bright spots showed that the person performed many activities, and the faint spots indicated that the person performed a small number of activities in space.

As shown in Figure 22, the person's behavior on the $11^{th}$, $12^{th}$, $13^{th,}$ and $14^{th}$ of April was similar to the person's behavior on the previous days, as there were two bright spots. There were also some minor differences these days, represented in the appearance of some simple activities in different spaces. On all days, there were two bright spots and some faint spots, except for the 14th of April, which contained only two bright spots, which indicated that the person's activity decreased and focused on two places only.

As shown in Figure 23, the behavior of the person on the 15th of April was similar to the behavior of the person on the previous days, and the activity of the person was changed on the $16^{th}$ and $17^{th}$ of April, which contained more activities in different spaces specially on the $17^{th}$ of April which contained several spots in many spaces. The person's behavior was decreased and focused in two places on the 19th of April.

As shown in Figure 24, the behavior of the person on the $20^{th}$ and $21^{st}$ increased and covered several spaces, and the activity of the person focused in the right direction of the image, which meant that the person spent most of his time in this space otherwise, the activity of the person decreased on $22^{nd}$ and $23^{rd}$ which meant that maybe the person spent most of his time outside the space.

As shown in Figure 25, the person's behavior on the 24th and 25th of April was similar to the behavior of the person on the first days that were mentioned, which contained two bright spots, and the person's behavior was focused on the right of the image. Compared with the $24^{th}$ and $25^{th}$ of April, the number of spots decreased on the $26^{th,}$ which meant that the activity of the person decreased in small spaces, but on the $27^{th}$, the activity of the person increased compared with the $26^{th}$ and the highest activity was found on the image's right.

As shown in Figure 26, the person's behavior on the 28th concentrated on four spots, and the highest activity was observed on the right of the image, but the person's activity on the $29^{th}$ decreased and focused on two spaces. The behavior of the person on the $30^{th}$ was similar to the behavior of the person on the $28^{th}$, and the behavior of the person on the 1st of May was identical to the behavior of the person on the $28^{th}$ and $30^{th,}$ but on the 1st of May, there were new spots at the top of the image which represented the presence of new activity for the person.

As shown in Figure 27, the person's behavior on the 2nd of May was similar to that of the person on the 1st of May. However, on the 2nd of May, there were new spots at the bottom of the image, which represented the presence of new activity for the person, and the behavior of the person on the 3rd of May is similar to the behavior of the person on the 2nd of May, but on the 3rd of May, the spots were more significant than the spots on the 2nd of May which meant that the number of activity of the person increased. The behavior of the person on the 5th of May was different from the previous days, which contained only one bright spot, which meant that the behavior decreased, and the behavior of the person on the 6th of May was like the behavior of the person on the $2^{nd}$ and $3^{rd}$ May.

In conclusion, there was an apparent reflection between spatial location and the number of hours of sleeping Activity and the number of hours of daily activity; for example, as shown in Figure 24, on the 20th of April, the activity of the person was higher in spatial location on the left hand sided which represented the sleeping activity which was reflected with the results for the number of sleeping activity as shown in Figure 18 and the activity of the person decreased in spatial location on the right-hand side which represented the daily activity which was reflected with the results for the number of daily activity and the number of hours of the daily activity was less than the number of hours of sleeping activity which was reflected in spatial location.

.

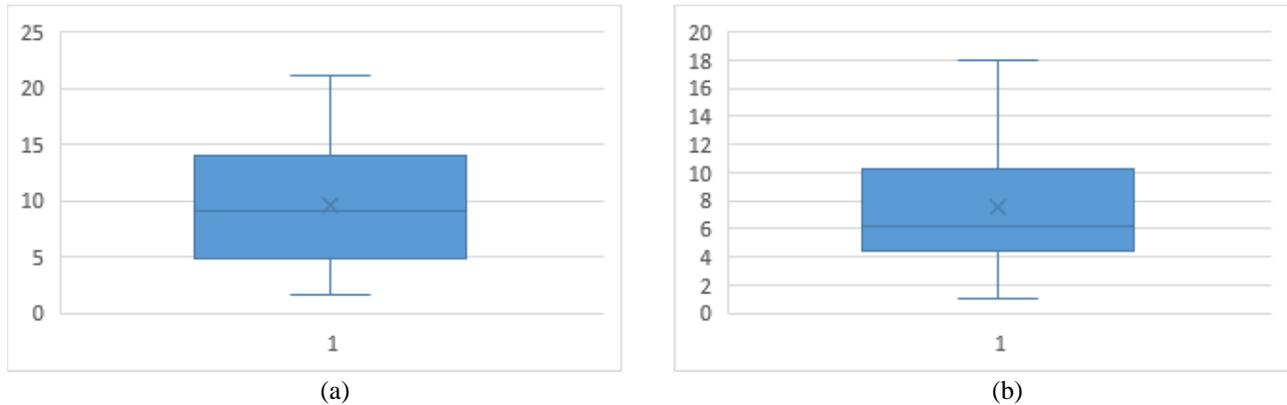

**Figure 29.** (a) Box plot for the number of hours for sleeping activity, and (b) Box plot for the number of hours for daily activity

## VI. CONCLUSION

Thermal sensor array (TSA) is very effective in monitoring human behavior because it overcomes problems related to privacy and high cost and determines the spatial location of the person to be monitored. It qualifies for various indoor applications, including monitoring energy efficiency, home security, and health. In this study, the activities were classified into three classes: sleeping activity, daily activity, and no activity. The study focused on monitoring sleeping activity and calculated the hours per day. The raw data was obtained in about 35 days, and the days when much data was missed were excluded. The percentage of missing data per day equals 20%. The individual sleeping and daily activities per day were represented. Results had shown that the person slept 9 hours per day on average, and his daily life activity was 7 hours per day on average. Bivariate distribution was applied to represent the spatial distribution of the person's activity. The spatial distribution was essential to show the relation between the person's activity and spatial location and describe the correlation between the activity and location. If the spatial distribution pattern changes, the person's behavior will change.

Results were compared with those of another study to calculate the hours spent sleeping and doing daily activities. Future work can be undertaken to do experiments to monitor all human activities by using the fusion of multiple thermal sensor arrays to cover a wide inspection area.

**Acknowledgments**
The authors would like to thank AmI-Lab members for providing the data used in the research, especially Prof. Bessam Abdulrazak.

**Author contributions**
Dina E. Abdelaleem was involved in data curation, visualization, and writing original draft preparation. Hassan M. Ahmed was involved in conceptualizing and writing the original draft preparation. M. Sami Soliman was involved in data curation, visualization, reviewing, and editing. Tarek M. Said was involved in data curation, visualization, reviewing, and editing. All authors read and approved the final manuscript.

**Funding**
Not Applicable.

**Data availability**
The datasets presented and analyzed during the current study are available from the corresponding author upon reasonable request.

**Declarations**

**Ethics approval and consent to participate.**
Not Applicable.

**Consent for publication**
Not Applicable.

**Competing interests**
The authors declare that they have no competing interests.

**REFERENCES**

[1] P. Chase, "Comment on 'The Origin of Modern Human Behavior,'" *Curr. Anthropol.*, vol. 44, no. 5, p. 637, 2003.

[2] M. Al-Khafajiy *et al.*, "Remote health monitoring of elderly through wearable sensors," *Multimed. Tools Appl.*, vol. 78, no. 17, pp. 24681–24706, 2019.

[3] B. Abdulrazak, H. Mostafa Ahmed, H. Aloulou, M. Mokhtari, and F. G. Blanchet, "IoT in medical diagnosis: detecting excretory functional disorders for Older adults via bathroom activity change using unobtrusive IoT technology," *Front. Public Heal.*, vol. 11, p. 1161943, 2023.

[4] H. M. Ahmed *et al.*, "Meal-time Detection by Means of Long Periods Blood Glucose Level Monitoring via IoT Technology," *arXiv Prepr. arXiv2303.00223*, 2023.

[5] A. Lotfi, C. Langensiepen, S. M. Mahmoud, and M. J. Akhlaghinia, "Smart homes for the elderly dementia sufferers: Identification and prediction of abnormal behaviour," *J. Ambient Intell. Humaniz. Comput.*, vol. 3, no. 3, pp. 205–218, 2012, doi: 10.1007/s12652-010-0043-x.

[6] S. Chandra Mukhopadhyay, "Wearable Sensors for Human Activity Monitoring," *IEEE Sens. J.*, vol. 15, no. 3, pp. 1321–1330, 2015.

[7] H. Cui and N. Dahnoun, "High precision human detection and tracking using millimeter-wave radars," *IEEE Aerosp. Electron. Syst. Mag.*, vol. 36, no. 1, pp. 22–32, 2021, doi: 10.1109/MAES.2020.3021322.

[8] A. Naser, A. Lotfi, J. Zhong, and J. He, "Heat-map based occupancy estimation using adaptive boosting," in *2020 IEEE International Conference on Fuzzy Systems (FUZZ-IEEE)*, 2020, pp. 1–7.

[9] Z. Chen and Y. Wang, "Remote Recognition of In-Bed Postures Using a Thermopile Array Sensor with Machine Learning," *IEEE Sens. J.*, vol. 21, no. 9, pp. 10428–10436, 2021, doi: 10.1109/JSEN.2021.3059681.

[10] A. Naser, A. Lotfi, and J. Zhong, "Towards human distance estimation using a thermal sensor array," *Neural Comput. Appl.*, vol. 2, 2021, doi: 10.1007/s00521-021-06193-2.

[11] V. Manuel Ionescu and F. Magda Enescu, "Low cost thermal sensor array for wide area monitoring," *Proc. 12th Int. Conf. Electron. Comput. Artif. Intell. ECAI 2020*, 2020, doi: 10.1109/ECAI50035.2020.9223193.

[12] A. Gomez, F. Conti, and L. Benini, "Thermal image-based CNN's for ultra-low power people recognition," in *Proceedings of the 15th ACM International Conference on Computing Frontiers*, 2018, pp. 326–331.

[13] J. Kemper and D. Hauschildt, "Passive infrared localization with a probability hypothesis density filter," in *2010 7th Workshop on Positioning, Navigation and Communication*, 2010, pp. 68–76.

[14] Z. Chen, C. Jiang, and L. Xie, "Building occupancy estimation and detection: A review," *Energy Build.*, vol. 169, pp. 260–270, 2018.

[15] H. Saha, A. R. Florita, G. P. Henze, and S. Sarkar, "Occupancy sensing in buildings: A review of data analytics approaches," *Energy Build.*, vol. 188, pp. 278–285, 2019.

[16] A. Howedi, A. Lotfi, and A. Pourabdollah, "Exploring entropy measurements to identify multi-occupancy in activities of daily living," *Entropy*, vol. 21, no. 4, 2019, doi: 10.3390/e21040416.

[17]	J. Yin, M. Fang, G. Mokhtari, and Q. Zhang, "Multi-resident location tracking in smart home through non-wearable unobtrusive sensors," *Lect. Notes Comput. Sci. (including Subser. Lect. Notes Artif. Intell. Lect. Notes Bioinformatics)*, vol. 9677, pp. 3–13, 2016, doi: 10.1007/978-3-319-39601-9_1.

[18]	Q. Li *et al.*, "Multi-resident type recognition based on ambient sensors activity," *Futur. Gener. Comput. Syst.*, vol. 112, pp. 108–115, 2020, doi: 10.1016/j.future.2020.04.039.

[19]	A. S. Crandall and D. J. Cook, "Coping with multiple residents in a smart environment," *J. Ambient Intell. Smart Environ.*, vol. 1, no. 4, pp. 323–334, 2009, doi: 10.3233/AIS-2009-0041.

[20]	A. R. Pratama, A. Lazovik, and M. Aiello, "Office Multi-Occupancy Detection using BLE Beacons and Power Meters," *2019 IEEE 10th Annu. Ubiquitous Comput. Electron. Mob. Commun. Conf. UEMCON 2019*, pp. 0440–0448, 2019, doi: 10.1109/UEMCON47517.2019.8993008.

[21]	A. Beltran, V. L. Erickson, and A. E. Cerpa, "ThermoSense: Occupancy thermal based sensing for HVAC control," *BuildSys 2013 - Proc. 5th ACM Work. Embed. Syst. Energy-Efficient Build.*, 2013, doi: 10.1145/2528282.2528301.

[22]	D. Qu, B. Yang, and N. Gu, "Infrared Physics & Technology Indoor multiple human targets localization and tracking using thermopile sensor," *Infrared Phys. Technol.*, vol. 97, no. November 2018, pp. 349–359, 2019, doi: 10.1016/j.infrared.2019.01.011.

[23]	M. Gochoo *et al.*, "Novel IoT-Based Privacy-Preserving Yoga Posture Recognition System Using Low-Resolution Infrared Sensors and Deep Learning," vol. 6, no. 4, pp. 7192–7200, 2019.

[24]	C. Zhong, W. W. Y. Ng, S. Zhang, C. Nugent, and C. Shewell, "Multi-occupancy Fall Detection using Non-Invasive Thermal Vision Sensor," vol. XX, no. XX, pp. 1–12, 2020, doi: 10.1109/JSEN.2020.3032728.

[25]	S. Savazzi, V. Rampa, L. Costa, and D. Tolochenko, "Processing of body-induced thermal signatures for physical distancing and temperature screening," vol. XX, no. XX, pp. 1–11, 2021, doi: 10.1109/JSEN.2020.3047143.

[26]	R. K. Rajeesh, A. M, B. E, S. J. P. J, K. A, and P. S, "Detection and monitoring of the asymptotic COVID-19 patients using IoT devices and sensors," *Int. J. Pervasive Comput. Commun.*, vol. 18, no. 4, pp. 407–418, 2022, doi: 10.1108/IJPCC-08-2020-0107.

[27]	E. Moisello, P. Malcovati, and E. Bonizzoni, "Thermal sensors for contactless temperature measurements, occupancy detection, and automatic operation of appliances during the covid-19 pandemic: A review," *Micromachines*, vol. 12, no. 2, pp. 1–20, 2021, doi: 10.3390/mi12020148.

[28]	A. Majeed and S. O. Hwang, "A Comprehensive Analysis of Privacy Protection Techniques Developed for COVID-19 Pandemic," *IEEE Access*, vol. 9, pp. 164159–164187, 2021, doi: 10.1109/ACCESS.2021.3130610.

[29]	G. Spasov, V. Tsvetkov, and G. Petrova, "Using IR array MLX90640 to build an IoT solution for ALL and security smart systems," *2019 IEEE XXVIII Int. Sci. Conf. Electron.*, pp. 1–4, 2019.

[30]	A. D. Shetty and W. S. Ab, "Detection and Tracking of a Human Using the Infrared Thermopile Array Sensor - ' Grid -EYE ,'" pp. 1490–1495, 2017.

[31]	M. Maaspuro, "A Low-Resolution IR-Array as a Doorway Occupancy Counter in a Smart Building," pp. 10–15.

[32]	V. Chidurala, G. S. Member, X. Li, and S. Member, "Occupancy Estimation Using Thermal Imaging Sensors and Machine Learning Algorithms," vol. 21, no. 6, pp. 8627–8638, 2021.

[33]	E. Song, H. Lee, J. Choi, and S. Lee, "AHD : Thermal-Image Based Adaptive Hand Detection for Enhanced Tracking System," vol. 4, no. c, pp. 1–13, 2018, doi: 10.1109/ACCESS.2018.2810951.

[34]	M. Vandersteegen, W. Reusen, K. Van Beeck, and T. Goedem, "Low-latency hand gesture recognition with a low resolution thermal imager," pp. 440–449, 2020, doi: 10.1109/CVPRW50498.2020.00057.

[35]	M. Raghavendra, A. M. Bharath, V. Dhananjaya, and S. Nithyanandhan, "IOT BASED THERMAL SURVILLANCE AND SECURITY SYSTEM," vol. 1, no. 2, pp. 93–98, 2019.

[36]	A. Tyndall, R. Cardell-Oliver, and A. Keating, "Occupancy estimation using a low-pixel count thermal imager," *IEEE Sens. J.*, vol. 16, no. 10, pp. 3784–3791, 2016.

[37]	G. Cerutti, R. Prasad, and E. Farella, "Convolutional neural network on embedded platform for people presence detection in low resolution thermal images," in *ICASSP 2019-2019 IEEE International Conference on Acoustics, Speech and Signal Processing (ICASSP)*, 2019, pp. 7610–7614.

[38]	M. Cokbas, P. Ishwar, and J. Konrad, "Low-resolution overhead thermal tripwire for occupancy estimation," in *Proceedings of the IEEE/CVF Conference on Computer Vision and Pattern Recognition Workshops*, 2020, pp. 88–89.

[39]	H. Mohammadmoradi, S. Munir, O. Gnawali, and C. Shelton, "Measuring people-flow through doorways using easy-to-install IR array sensors," *Proc. - 2017 13th Int. Conf. Distrib. Comput. Sens. Syst. DCOSS 2017*, vol. 2018-Janua, pp. 35–43, 2018, doi: 10.1109/DCOSS.2017.26.

[40] H. Aloulou, M. Mokhtari, T. Tiberghien, J. Biswas, and P. Yap, "An adaptable and flexible framework for assistive living of cognitively impaired people," *IEEE J. Biomed. Heal. informatics*, vol. 18, no. 1, pp. 353–360, 2013.

[41] A. Naser, A. Lotfi, and J. Zhong, "A Novel Privacy-Preserving Approach for Physical Distancing Measurement Using Thermal Sensor Array," *ACM Int. Conf. Proceeding Ser.*, pp. 81–85, 2021, doi: 10.1145/3453892.3453903.

[42] A. Naser, A. Lotfi, and J. Zhong, "Adaptive Thermal Sensor Array Placement for Human Segmentation and Occupancy Estimation," *IEEE Sens. J.*, vol. 21, no. 2, pp. 1993–2002, 2021, doi: 10.1109/JSEN.2020.3020401.

[43] H. M. Ahmed and B. Abdulrazak, "Monitoring Indoor Activity of Daily Living using Thermal Imaging: A Case Study," *Int. J. Adv. Comput. Sci. Appl.*, vol. 12, no. 9, pp. 11–16, 2021, doi: 10.14569/IJACSA.2021.0120902.

[44] H. M. Ahmed, S. Maraoui, B. Abdulrazak, B. Cossette, and F. G. Blanchet, "Drug Intervention Follow up with Internet of Things: A Case Study," in *International Conference on Smart Homes and Health Telematics*, 2023, pp. 65–75.

[45] A. Naser, A. Lotfi, and J. Zhong, "Multiple Thermal Sensor Array Fusion Toward Enabling Privacy-Preserving Human Monitoring Applications," vol. 9, no. 17, pp. 16677–16688, 2022.

[46] H. M. Ahmed, B. Abdulrazak, F. G. Blanchet, H. Aloulou, and M. Mokhtari, "Long Gaps Missing IoT Sensors Time Series Data Imputation: A Bayesian Gaussian Approach," *IEEE Access*, 2022.